\documentclass[lettersize,journal]{IEEEtran}
\usepackage{cite}
\usepackage{amsmath,amssymb,amsfonts}
\usepackage{fancyhdr} % Ò³Ã¼Ò³½Å
\usepackage{algorithmic}
\usepackage{graphicx}
\usepackage{booktabs}
\usepackage{textcomp}
\usepackage{subfigure}
\usepackage{bm}
\usepackage{algorithm}
\usepackage{array}
\usepackage[caption=false,font=normalsize,labelfont=sf,textfont=sf]{subfig}
\usepackage{stfloats}
\usepackage{url}
\usepackage{verbatim}

\newtheorem{definition}{$\mathbf{Definition}$}
\newtheorem{theorem}{$\mathbf{Theorem}$}

\newtheorem{problem}{$\mathbf{Problem}$}
\newtheorem{example}{$\mathbf{Example}$}
\begin{document}
\title{Flexible Active Safety Motion Control for Robotic Obstacle Avoidance: A CBF-Guided MPC Approach} 
%\title{FAST: Flexible Active Safety Tracking Control for Robot Manipulators Utilizing CBF-Guided MPC Framework} 
\author{Jinhao Liu, Jun Yang, Jianliang Mao, Tianqi Zhu, Qihang Xie, Yimeng Li, Xiangyu Wang, Shihua Li
	\thanks{Jinhao Liu, Tianqi Zhu, Yimeng Li, Xiangyu Wang and Shihua Li are with the School of Automation and the Key Laboratory of Measurement and Control of Complex Systems of Engineering, Ministry of Education, Southeast University, Nanjing 210096, China (email: jinhaoliu@seu.edu.cn; tianqi$\_$zhu@seu.edu.cn; lym$\_$seu@seu.edu.cn; w.x.y@seu.edu.cn; lsh@seu.edu.cn).}
	\thanks{Jun Yang is with Department of Aeronautical and Automotive Engineering, Loughborough University, Loughborough LE11 3TU, U.K. (e-mail: j.yang3@lboro.ac.uk).}
	\thanks{Jianliang Mao and Qihang Xie are with the College of Automation Engineering, Shanghai University of Electric Power, Shanghai 200090, China (e-mail: jl$\_$mao@shiep.edu.cn; q.h.xie@mail.shiep.edu.cn).}}
\markboth{}%
{Shell \MakeLowercase{\textit{et al.}}: A Sample Article Using IEEEtran.cls for IEEE Journals}

\maketitle
\begin{abstract}
A flexible active safety motion (FASM) control approach is proposed for the avoidance of dynamic obstacles and the reference tracking in robot manipulators. The distinctive feature of the proposed method lies in its utilization of control barrier functions (CBF) to design flexible CBF-guided safety criteria (CBFSC) with dynamically optimized decay rates, thereby offering flexibility and active safety for robot manipulators in dynamic environments. First, discrete-time CBFs are employed to formulate the novel flexible CBFSC with dynamic decay rates for robot manipulators. Following that, the model predictive control (MPC) philosophy is applied, integrating flexible CBFSC as safety constraints into the receding-horizon optimization problem. Significantly, the decay rates of the designed CBFSC are incorporated as decision variables in the optimization problem, facilitating the dynamic enhancement of flexibility during the obstacle avoidance process. In particular, a novel cost function that integrates a penalty term is designed to dynamically adjust the safety margins of the CBFSC. Finally, experiments are conducted in various scenarios using a Universal Robots 5 (UR5) manipulator to validate the effectiveness of the proposed approach.
\end{abstract}
\vspace{-0mm}
\section{Introduction}
\label{sec:introduction}
\IEEEPARstart{R}{obot} manipulators are extensively utilized in manufacturing and assembly lines in various industrial applications, including the automotive industry \cite{application1}, medicine \cite{application2}, and aerospace \cite{application3}, due to their ability to replace or assist humans in completing some complex and repetitive tasks \cite{robot_uncer}. Beyond accomplishing tracking tasks, guaranteeing safety is a paramount requirement for robot manipulators. However, their operational workspace often involves interaction with both humans and dynamic environments (typically characterized by the presence of moving obstacles), inevitably raising safety concerns \cite{intro1}. Confronting these challenges requires the development of active safety control strategies for robot manipulators to guarantee critical safety in dynamic environments.
\subsection{Related Work}
One fundamental concern in ensuring the operational safety of robot manipulators is to address the challenge of avoiding obstacles. Methods for tackling obstacle avoidance can typically be categorized into two categories: offline global path planning and online local path planning \cite{plan1}. 

For static environments/obstacles, offline path planning can achieve obstacle avoidance. Typical offline path planning algorithms for autonomous robots can be categorized into the following types: sampling-based methods \cite{RRT1,RRT2}, search-based methods \cite{search_based1,search_based2}, heuristic-based methods \cite{H2,H3}, 
and optimization-based methods \cite{Op1}. However, when addressing path-planning challenges in dynamic environments, particularly those with fast-moving obstacles, offline path planning methods are not suitable. Therefore, it is necessary to incorporate a local real-time planner to refine the global path and adapt to dynamic environments.

For dynamic environments/obstacles, the artificial potential field (APF) method stands as one of the typical online obstacle avoidance algorithms. It operates by establishing a global attractive field around the target and a local repulsive field around obstacles. By integrating the effects of both fields, the robot can navigate towards the target while effectively avoiding obstacles \cite{APF1,APF2,mpc_safe3}. Fuzzy logic approaches determine suitable actions to avoid dynamic obstacles by employing fuzzy sets and rules inspired by human decision-making \cite{fuzzy1, fuzzy2}. Reinforcement learning (RL) methods commonly employ reward functions specifically designed for dynamic obstacle avoidance. Through RL algorithms, robot manipulators learn from previous experience to discern actions that optimize reward functions, thereby achieving effective obstacle avoidance \cite{learn1,learn2}.
%Jacobian pseudoinverse-based methods However, these methods need to calculate the pseudoinverse of the Jacobian matrix in real-time, which is time consuming and leads to a decline of planning performance, as well as even posing a security risk
Note that the above-mentioned approaches lack the capability to seamlessly integrate online
path planning, tracking control, and constraints handling within a unified framework, thus failing to address the discrepancy between the dynamic nature of the environments and the relatively sluggish nature of the higher-level planner. Unlike the aforementioned methods, inequality-based approaches \cite{qpnn,QP1,QP2} serve as effective tools to achieve dynamic obstacle avoidance by incorporating obstacle avoidance tasks as inequality constraints into the optimization problem. Model Predictive Control (MPC), which dynamically solves finite-horizon optimal control problems by fully utilizing system models, has attracted significant attention due to its efficiency in handling various constraints \cite{ mpc3}. MPC methods are popular for addressing dynamic obstacle avoidance problems due to their ability to solve constrained optimization problems that allow path planning, tracking control, and constraints handling to be integrated into a single framework. Typical applications of MPC methods to tackle dynamic obstacle avoidance problems for robot manipulators have been investigated in \cite{mpc_safe3,mpc_safe1,mpc_safe2,mpc_safe4,mpc_safe5}. Although most of the methods mentioned above rely on simple safety criteria defined by Euclidean norms, which become active only when robots are in close proximity to obstacles \cite{mpccbf1}, these approaches can compromise safety in scenarios with fast-moving dynamic obstacles. 

Recently, control barrier functions (CBF) have emerged as powerful techniques to ensure forward invariance of a designated safe set \cite{intro1,CBF1,Dis_CBF,Dis_CBF2}. 
%A comprehensive survey covering both theoretical and practical aspects of CBFs in the field of robot manipulators can be found in \cite{intro1}. 
Several studies have used CBFs to ensure active safety of robot manipulators, as reported in \cite{CBF_rob1,CBF_rob2,CBF_rob5}. To be specific, \cite{CBF_rob1} demonstrated how to guarantee safety-critical kinematic constraints for robotic systems through CBF. \cite{CBF_rob2} combined CBFs for safety and control Lyapunov functions for task constraints in addressing obstacle avoidance for robot manipulators. 
%In \cite{CBF_rob3}, high-order CBFs were combined with computed torque control for UR-type manipulators to ensure safety while minimizing input changes. 
\cite{CBF_rob5} combined a learning-based MPC and a CBF-based safety filter for obstacle avoidance. However, these methods are solely focused on the avoidance of static obstacles. More importantly, they utilize CBFs with fixed decay rates for safety filter design, which lacks flexibility and potentially compromises controller feasibility, as discussed in \cite{CBF_fea1,CBF_fea2,CBF_fea3,CBF_fea4}. In particular, \cite{CBF_fea4} emphasized the MPC design with CBFs incorporating dynamic decay rates to ensure critical teleoperation safety. However, dynamic obstacle scenarios were not addressed in \cite{CBF_fea4}. 
\vspace{-3.8mm}
\subsection{Contributions}
This paper focuses primarily on online path planning and tracking control at the kinematic level, aiming to achieve flexible active safety motion (FASM) control for robot manipulators. We propose a flexible CBF-guided safety criteria (CBFSC) integrating dynamically optimized decay rates to enhance flexible active safety for dynamic obstacle avoidance. Subsequently, the MPC framework is applied to simultaneously realize online path planning, tracking control, and constraint satisfaction. Real-world experiments are conducted in dynamic scenarios using a Universal Robots 5 (UR5) manipulator to validate the effectiveness of the proposed method. The primary contributions of this paper are summarized below:
\begin{enumerate}[]       	
	\item A CBF-guided MPC framework is proposed to simultaneously achieve dynamic obstacle avoidance and reference tracking for robot manipulators. By integrating CBFSC within the MPC framework, the proposed approach offers active safety to avoid potential collisions at an earlier stage, while strictly guaranteeing constraints on control input.
	
	\item The proposed FASM control provides flexible safety strategies. First, the flexible CBFSC employs dynamically optimized decay rates to prioritize active safety when the robot manipulator is distant from obstacles. Then it gradually relaxes the safety restrictions as the robot manipulator approaches obstacles to improve the feasibility of optimization problems. Second, a penalty term is incorporated into the cost function to dynamically adjust the safety margins of the CBFSC. 
\end{enumerate}
\section{Preliminaries and Problem Formulation}
In this section, the preliminaries on robot kinematics and the concepts of discrete-time CBFs are first reviewed. The problem of dynamic obstacle avoidance, reference tracking, and constraint satisfaction for robot manipulators is then formulated.
\subsection{Robot Kinematics}
In this study, obstacle avoidance is implemented by identifying critical points on each link of the robot manipulator and ensuring that these critical points satisfy the specified safety conditions. Specifically, at time step $k$, let $x_{j,k} \in \mathbb{X}_j \subset \mathbb{R}^3$ denote the $j$th ($j \in \mathbb{N}_{\geq0}$) critical point position, and $x_{e,k}\in \mathbb{X}_e \subset \mathbb{R}^7$ denotes the end-effector pose. Subsequently, the discrete-time forward kinematics of a robot manipulator can be obtained through the Jacobian matrix as follows \cite{mpc_safe3}
\begin{equation}\label{jaco1}
	x_{j,k+1}   = x_{j,k} + t_s J_j(\theta_k)u_k,
\end{equation}
\begin{equation}\label{jaco2}
	x_{e,k+1}   = x_{e,k} + t_sJ_e(\theta_k)u_k,
\end{equation}
where $t_s$ is the sampling period, $u_k \in \mathbb{U} \subset \mathbb{R}^n$ is the joint velocity, which is indicated as system input, $n$ represents the number of degrees of freedom, $\theta_k \in \mathbb{R}^n$ is the joint position, $J_j(\theta_k)$ and $J_e(\theta_k)$ denote the Jacobian matrices of the $j$th critical point and the end-effector, respectively. In this paper, the design process of the proposed method will mainly focus on the end effector $x_{e,k}$. Since the design process for other critical points $x_{j,k}$ follows a similar methodology, their details are omitted to avoid repetition and maintain brevity.
\subsection{Discrete-time Control Barrier Functions}
First, we briefly review the basic results of discrete-time CBFs \cite{Dis_CBF,Dis_CBF2}. Consider a discrete-time control system as follows:
\begin{equation}\label{non}
	x_{k+1} = f(x_{k},u_k),
\end{equation}
where $x_k \in \mathbb{Q} \subset \mathbb{R}^q$ and $u_k \in U \subset \mathbb{R}^p$ represent the states and control inputs of the system, respectively. Then, define a safe set $\mathbb{C}$ for the system (\ref{non}) which is denoted as $\mathbb C:=\{x_k \in \mathbb{Q} \ | \ B(x_k) \geq 0 \}$, $\partial \mathbb C:=\{x_k \in \mathbb{Q} \ | \ B(x_k) = 0 \}$, where $B: \mathbb{Q} \rightarrow \mathbb R$ is a continuous function. The set $\mathbb{C}$ is forward invariant for the system (\ref{non}) if for any $x_0 \in \mathbb{C}$, $x_k \in \mathbb{C}$, $\forall k \in \mathbb{N}_{\geq0}$. Next, we show that the set $\mathbb{C}$ is forward invariant by using the exponential CBF formulation. 
\begin{figure}[t]
	\centering{\includegraphics[width=0.45\textwidth]{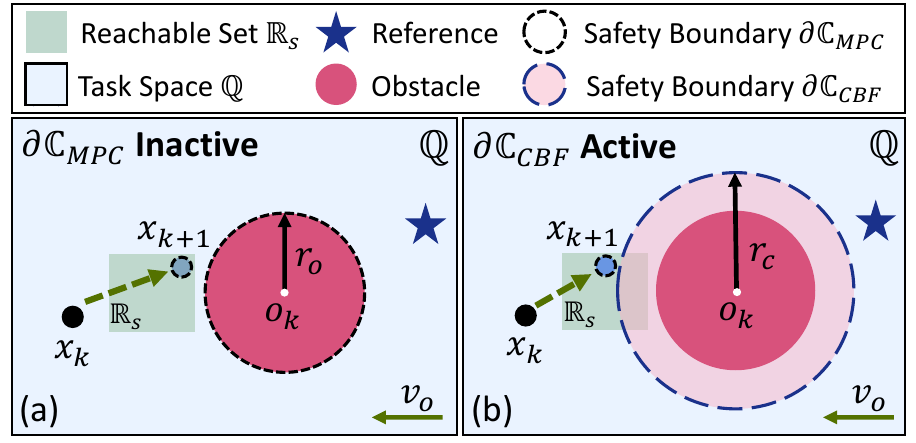}}
	\caption{The advantages of the CBFSC. (The next-step reachable set $\mathbb{R}_s := \{x_{k+1} \in \mathbb{Q} \ | \  x_{k+1} = f(x_{k},u_k), u_k \in U  \}$.)}
	\label{fig}
\end{figure}
\begin{definition}
	(Discrete-Time Exponential CBFs \cite{Dis_CBF}) A function $B:\mathbb{Q} \rightarrow \mathbb R$ is a discrete-time exponential CBF for the system (\ref{non}) if: $B(x_0) \geq 0$ and there exists a control input $u_k \in U$ such that 
	\begin{equation}
		\Delta B(x_{k},u_k) + \gamma B(x_{k}) \geq 0, \ \forall k \in \mathbb{N}_{\geq0}, \ 0 < \gamma \leq 1,
	\end{equation}
	where $\Delta B(x_{k},u_k) = B(x_{k+1}) - B(x_{k})$, $\gamma$ is the decay rate.
\end{definition}

Next, we will illustrate the inherent advantages of using the CBFSC for obstacle avoidance through a simple example.
\begin{example}
	Consider an obstacle avoidance scenario for the system (\ref{non}) in a two-dimensional plane as depicted in Fig. 1. Here, the moving obstacle is represented by a red circle centered at $o_{k}$ with a radius of $r_o$ and a velocity of $v_o$.
\end{example}
\begin{enumerate}[]	
	\item Under MPC methods, the safety constraint is defined as $S(x_{k}) = \|x_{k} -  o_{k}\| - {r_{o}} \geq 0$, indicating $x_{k+1} \geq r_o + o_{k+1}$ at time step $k+1$ (consider only the simple case where $x_{k}\geq0$, $o_{k}\geq0$). The safe boundary is denoted as $\partial {\mathbb C}_{MPC}:=\{x_{k+1} \in \mathbb{Q} \ | \ S(x_{k+1}) = 0 \}$.
	
	\item Differently, under the CBFSC, the safe constraint is expressed as $	S(x_{k+1}) \geq (1- \gamma)S(x_{k})$, with $0 < \gamma \leq 1$. This implies $x_{k+1} \geq r_c + o_{k+1}$ at time step $k+1$, where $r_c = \sqrt{(1- \gamma)S(x_{k}) + {r_o}^2}$. The safe boundary satisfying the CBFSC is defined as $\partial {\mathbb C}_{CBF}:=\{x_{k+1} \in \mathbb{Q} \ | \ S(x_{k+1}) - (1- \gamma)S(x_{k})=0, \ 0 < \gamma \leq 1 \}$.
\end{enumerate}

As illustrated in Fig. 1(a), the safety criteria in traditional MPC methods remain inactive since the current state $x_k$ is far from the obstacle, resulting in no intersection between $\mathbb{R}_s$ and $\partial \mathbb C_{MPC}$. Consequently, the system does not trigger avoidance actions at time step $k$, leading to $x_{k+1}$ approaching the obstacle closely, which poses a safety risk, especially in cases involving fast-moving obstacles. Moreover, since the decay rate $\gamma$ in the CBFSC can be adjusted within the range of $0 < \gamma \leq 1$, it ensures that $r_c \geq r_o$, thus enhancing the possibility of achieving the intersection of $\partial \mathbb C_{CBF}$ with $\mathbb{R}_s$, as shown in Fig. 1(b). Consequently, even when $x_k$ is far from the obstacle, the CBFSC can trigger avoidance actions by selecting $\gamma$ to a small value. This active safety nature is particularly beneficial in scenarios with fast-moving obstacles, initiating avoidance actions at an earlier stage. However, most CBF methods for obstacle avoidance utilize fixed decay rates $\gamma$ for the design of safety criteria, lacking flexibility and potentially compromising the feasibility of the controller. Therefore, in this paper, we will propose a flexible CBFSC that incorporates dynamically optimized decay rates $\gamma_k$.
\vspace{-8mm}
\subsection{Problem Formulation}
The primary control objective of this study is to ensure that the end effector accurately tracks a desired position and pose, while actively avoiding dynamic obstacles along its path and simultaneously satisfying the control constraints. 

To quantify the tracking objective, let $s_k\in \mathbb{X}_e \subset \mathbb{R}^7$ be the desired position and pose of the end effector. Next, the tracking error is defined as $e_{k}   = s_k - x_{e,k}$. In this paper, we assume that the obstacles are spheres, where $o_k$ represents the spatial position of the obstacle with a radius of $R_o$. To meet the demands of dynamic obstacle avoidance, the safe sets for systems (\ref{jaco1}) and (\ref{jaco2}) are defined as follows
\begin{equation}\label{safe_set1}
	\mathbb S_e:=\{x_{e,k} \in \mathbb X_e \ | \ h_{e,k} \geq 0 \},
	\mathbb S_j:=\{x_{j,k} \in \mathbb X_j \ | \ h_{j,k} \geq 0 \}, 
\end{equation}
where $h_{e,k}   = \|x_{e,k}-o_k\|-d_{\rm min}-R_o$, $h_{j,k}   = \|x_{j,k}-o_k\|-d_{\rm min}-R_o$, $d_{\rm min}$ represents the minimum safe distance between the robot manipulator and the obstacle. Thus, the control problem studied in this paper is formulated as follows.
\begin{problem}
	Consider the robot kinematics described by (\ref{jaco1}) and (\ref{jaco2}), design a control law $u_k\in \mathbb{U}$ to drive the tracking error $e_k$ to zero while guaranteeing the robot manipulator in the safe set, that is,
	\begin{equation}\label{problem}
			\mathop {\lim }\limits_{k \to \infty } e_{k} = 0,\ u_k\in \mathbb{U},\ x_{e,k} \in \mathbb{S}_e,\ x_{j,k} \in \mathbb{S}_j.
	\end{equation}
\end{problem}
\section{Main Results}
In this section, we propose a novel CBF-guided MPC solution to achieve FASM control for robot manipulators. First, a GPIO is introduced to estimate the dynamic information of the obstacle. Then, the design of flexible CBFSC for dynamic obstacle avoidance is explained. Finally, we present the FASM controller design. The detailed design of each module will be presented step by step.
\subsection{Observer Design for Dynamic Information of Obstacles}
At time step $k$, we define $o_k = \left[ {\begin{array}{*{20}{c}} {o_{x,k}}&{o_{y,k}}&{o_{z,k}}\end{array}} \right]^{\rm T}$. The higher-order information of the obstacle is denoted as $D_k = \left[ {\begin{array}{*{20}{c}}{o_{k}}&{l_{1,k}}&{\cdots}&{l_{m-1,k}}\end{array}} \right]^{\rm T}$, ${l_{1,k}}$ to ${l_{m-1,k}}$ are the higher-order differences of $o_k$. It is supposed that ${l_{m,k}}=0$, with $m$ as the order of $o_k$. In particular, ${l_{1,k}}$ denotes the obstacle velocity $v_k = \left[ {\begin{array}{*{20}{c}} {v_{x,k}}&{v_{y,k}}&{v_{z,k}}\end{array}} \right]^{\rm T}$. Then, $D_k$ and $o_k$ are assumed to be generated by $D_{k+1} = A D_k$, $o_{k} = {C}D_k$, 
where ${A} =  \left[ {\begin{array}{*{20}{c}}
		1&{{t_s}}& \cdots &0\\
		0&1& \ddots & \vdots \\
		\vdots & \ddots & \ddots &{{t_s}}\\
		0& \cdots &0&1
\end{array}} \right]_{m \times m}$, $C= \begin{bmatrix}
I & 0 & {\cdots} & 0\end{bmatrix}_{1 \times m}$.

In practice, $D_k$ is possibly not exactly known, that is, only the spatial position $o_k$ is known. For this case, a GPIO is designed to estimate $D_k$ as follows
\begin{equation}\label{gpio2}
	\left\{ \begin{matrix}
		\begin{aligned}
			\displaystyle  \xi_{1,k+1} &= \xi_{1,k} + t_s[\xi_{2,k} + \alpha_1 (o_k-\xi_{1,k})],\\
			\displaystyle  \xi_{2,k+1} &= \xi_{2,k} + t_s[\xi_{3,k} + \alpha_2 (o_k-\xi_{1,k})],\\
			\displaystyle	\vdots \ &  \\
			\displaystyle \xi_{m,k+1} &= \xi_{m,k} + t_s\alpha_m (o_k-\xi_{1,k}),
		\end{aligned}
	\end{matrix} \right.
\end{equation}
where $\xi_{1,k}$ is the estimation of $o_k$, denotes as $\xi_{1,k} = \hat o_k$, $\xi_{2,k} = \hat v_k$, $\xi_{i,k}= \hat l_{i-1,k}$, $i \in \mathbb{N}_{3:m}$. Define the errors of the observer as $\epsilon_{1,k} = o_{k} - \xi_{1,k}$, $\epsilon_{i,k}= l_{i-1,k} - \xi_{i,k}$, $i \in \mathbb{N}_{2:m}$. The estimation error system of the observer (\ref{gpio2}) is defined as
\begin{equation}\label{obser_error2}
	E_{k+1} = {\left[ {\begin{array}{*{20}{c}}
				{\epsilon_{1,k+1}}&{\epsilon_{2,k+1}}&{\cdots}&{\epsilon_{m,k+1}}
		\end{array}} \right]^{\rm T}}=\Phi E_{k}, 
\end{equation}
where $\Phi = \left[ {\begin{array}{*{20}{c}}
		{1 - {t_s}{\alpha _1}}&{{t_s}}&0&\ldots &0\\
		{ - {t_s}{\alpha  _2}}&1&{{t_s}}&\ldots &0\\
		\vdots & \vdots & \ddots & \ddots & \vdots\\
		{ - {t_s}{\alpha  _{m - 1}}}&0&\ldots&1&{{t_s}}\\
		{ - {t_s}{\alpha _m}}&0&\ldots&0&1
\end{array}} \right]_{m \times m}$. 

Suppose that the error system \eqref{obser_error2} is asymptotically stable by designing $\alpha_i$ to satisfy $\rho(\Phi)<1$, where $\rho(\Phi)<1$ denotes
the spectral radius of the matrix $\Phi$. 

Now, we are in a position to quantify the estimation errors. First, define $V(E_k) = \|E_k\|_{W}^2$, $W$ is the solution to the linear matrix inequality ${\Phi^{\rm T}} W \Phi - \eta^2 W \leq 0$ with $\rho(\Phi) < \eta <1$. Given $c_1 = \lambda_{\rm min} (W)$ and $c_2 = \lambda_{\rm max} (W)$, which yield
\begin{equation}
	c_1 \|E_k\|^2 \leq V(E_k) \leq c_2 \|E_k\|^2,
\end{equation}
\begin{equation}
	V(E_{k+1}) = \|E_k\|_{{\Phi^{\rm T}}W\Phi}^2 \leq \|E_k\|_{\eta^2 W}^2 \leq \eta^2 V(E_{k}).
\end{equation}
Now, we have
\begin{equation}\label{ineq}
	\|E_k\|^2 \leq \dfrac{V(E_k)}{c_1} \leq \dfrac{ \eta^{2k}}{c_1} V(E_0) \leq \dfrac{c_2 \eta^{2k}}{c_1}\|E_0\|^2 \leq {\phi_k}^2 \delta^2,
\end{equation}
where $\phi_k = \eta^{k}\sqrt{\dfrac{c_2}{c_1}}$, $E_0$ is the initial estimation error, $\delta$ is assumed to be the known upper bound of $\|E_{0}\|$.
\subsection{Flexible CBFSC for Obstacle Avoidance}
In this subsection, we first design the new safety distance considering the estimation errors, and then we proceed to design the flexible CBFSC with dynamic decay rates.

The estimation errors of obstacle information may impact the safety of obstacle avoidance for robot manipulators. Therefore, we consider utilizing the quantification of the estimation error (\ref{ineq}) to design the new safety distance as
\begin{equation}\label{r_safe}
	r_{{\rm safe}}  = d_{\rm min} + R_o + r_{d}, 
\end{equation}
where $r_{d} =\delta \phi_0$ represents the tolerance distance to counteract the effects of estimation errors. Based on the definition of $r_{{\rm safe}}$, we can further define a function to describe the surplus safety distance between the end-effector and the obstacle as
\begin{equation}\label{B}
	H(x_{e,k}, o_{k}) =  \|x_{e,k} -  o_{k}\| - {r_{{\rm safe}}}.
\end{equation}

Next, we will fully exploit the concept of discrete-time CBFs introduced by Definition 1 to design the CBFSC, ensuring that $H(x_{e,k}, o_{k})$ behaves as an exponential CBF for the end-effector (\ref{jaco2}), thereby maintaining the the forward invariance of the safe set $\mathbb S_e$.
\begin{theorem}\label{thm1}
At the time step $k$, based on Definition 1, consider the end effector (\ref{jaco2}), the safe set $\mathbb{S}_e$ and the observer (\ref{gpio2}). Given $x_{e,k} \in \mathbb S_e$, $H(x_{e,k}, o_{k}) \geq 0$, the proposed flexible CBFSC is designed as
\begin{equation}\label{CBF}
	\Delta H(x_{e,k}, u_k, o_{k}) +  \gamma_k H( x_{e,k}, o_{k}) \geq 0, \ 0 < \gamma_k \leq 1,
\end{equation}
where $\Delta H(x_{e,k}, u_k, o_{k}) = H(x_{e,k+1}, \hat o_{k+1})-H( x_{e,k}, o_{k})$, $\gamma_k$ is a variable to be optimized. Any feasible controller $u_k$ that satisfies the CBFSC (\ref{CBF}) guarantees that the safe set $\mathbb{S}_e$ defined in (\ref{safe_set1}) is forward invariant for the end effector (\ref{jaco2}).
\end{theorem} 
\begin{IEEEproof}
	Keep in mind that $o_{k+1} - \hat o_{k+1}  = \epsilon_{1,k+1}$ according to (\ref{gpio2}) and (\ref{obser_error2}). Based on the observer (\ref{gpio2}), the safety distance (\ref{r_safe}), and the definition (\ref{B}), we have the following.	
	\begin{equation}\label{eqe}
		\begin{aligned}
			H(x_{e,k+1}, \hat o_{k+1}) = & \ \|x_{e,k+1} -  \hat o_{k+1}\| - {r_{{\rm safe}}}\\
			= &\ \|x_{e,k+1} - o_{k+1} + \epsilon_{1,k+1} \|  - {r_{{\rm safe}}} \\
%			\leq &\  \|x_{e,k+1} - o_{k+1}\| + \| \epsilon_{1,k+1}  \|  - {r_{{\rm safe},k+1}} \\
			\leq & \  \|x_{e,k+1} - o_{k+1}\| - d_{\rm min} - R_o \\
			&+  \|\epsilon_{1,k+1}  \|-  \delta \phi_{0}.
		\end{aligned}
	\end{equation}

	According to (\ref{ineq}), one has $\| \epsilon_{1,k+1} \|=\| C E_{k+1} \| \leq \| C\| \| E_{k+1} \| \leq \delta \phi_{k+1}\leq \delta \phi_{0}$. Now, we can rewrite (\ref{eqe}) as
	\begin{equation}\label{eqe2}
		\|x_{e,k+1} - o_{k+1}\| - d_{\rm min} - R_o \geq H(x_{e,k+1}, \hat o_{k+1}).
	\end{equation}
	Keep in mind that $x_{e,k} \in \mathbb S_e$ and $H(x_{e,k}, o_{k}) \geq 0$, the CBFSC (\ref{CBF}) implies that $H(x_{e, k+1}, \hat o_{k+1}) \geq (1-\gamma_k) H( x_{e,k}, o_{k}) \geq 0, \ \forall k \in \mathbb{N}_{\geq0}$. Thus, according to (\ref{safe_set1}) and (\ref{eqe2}), we have that $h_{e,k+1}   = \|x_{e,k+1}-o_{k+1}\|-d_{\rm min} - R_o \geq 0$ holds for $\forall k \in \mathbb{N}_{\geq0}$, which implies $x_{e,k+1} \in \mathbb S_e$, $\forall k \in \mathbb{N}_{\geq0}$.
\end{IEEEproof}

The design progress of the CBFSC for the critical points $x_{j,k}$ is similar to that of the end effector. Therefore, we omit it here and present the specific conclusion directly. The CBFSC for $x_{j,k}$ are designed as follows
\begin{equation}\label{CBF2}
	\Delta H(x_{j,k}, u_k, o_{k}) +  \gamma_{j,k} H( x_{j,k}, o_{k}) \geq 0, \ 0 < \gamma_{j,k} \leq 1,
\end{equation}
where $\Delta H(x_{j,k}, u_k, o_{k}) = H(x_{j,k+1}, \hat o_{k+1})-H( x_{j,k}, o_{k})$. 

Now, we can utilize CBFSC (\ref{CBF}) and (\ref{CBF2}) as safety constraints within the MPC framework to regulate the $u_k$ of robot manipulators.
\subsection{FASM Controller Design}
First, considering the following stage cost function
\begin{equation}\label{sc}
	\begin{aligned}
		F_{sc}(x_{e,k},u_{k},\gamma_{k},\gamma_{j,k}) = & \ \|x_{e,k} - s_k\|_{Q}^2 +\left\|  u_{k}\right\|_{R}^2 + \left\|  \gamma_{k}\right\|_{P_{\gamma}}^2 \\
		& + \left\|  \gamma_{j,k}\right\|_{P_{j}}^2,
	\end{aligned}
\end{equation}
where $Q$, $R$, ${P_j}$ and ${P_\gamma}$ are positive definite, $j \in \mathbb{N}_{\geq0}$. Parameters ${P_\gamma}$ and ${P_j}$ are integrated into the stage cost to adjust the overall values of ${\gamma}_{k}$ and ${\gamma}_{j,k}$, respectively, thus fine-tuning the safety margins of the CBFSC (\ref{CBF}) and (\ref{CBF2}).

At time step $k$, based on the systems (\ref{jaco1}), (\ref{jaco2}), the observer (\ref{gpio2}), the flexible CBFSC (\ref{CBF}), (\ref{CBF2}), and the stage cost (\ref{sc}), the MPC optimization problem for achieving FASM control for robot manipulators is formulated as follows
\begin{subequations}\label{op}
	\begin{align}
		&\mathop{\mathrm{min}}\limits_{\bm{{u}}_{k}, {\gamma}_{k}, {\gamma}_{j,k}}\: J_{k} =\sum\limits_{i = 0}^{ N}{F_{sc}(x_{e,i|k}, u_{i|k}, \gamma_{k}, \gamma_{j,k})}  \\
		s.t. \quad 
		&  x _{e,i+1|k} = x_{e,i|k} + t_s J_{e}(\theta_k)u_{i|k}, \ x_{e,0|k} = x_{e,k},\\
		&  x _{j,i+1|k} = x_{j,i|k} + t_s J_{j}(\theta_k)u_{i|k}, \ x_{j,0|k} = x_{j,k}, \\
		& \Delta H(x_{e,i|k}, u_{i|k}, o_{i|k}) +  \gamma_k H( x_{e,i|k}, o_{i|k}) \geq 0, \\
		& \Delta H(x_{j,i|k}, u_{i|k}, o_{i|k}) +  {\gamma}_{j,k} H( x_{j,i|k}, o_{i|k}) \geq 0, \\
		& \hat o_{i+1|k} = CA\hat D_{i|k}, \hat D_{i+1|k}=A\hat D_{i|k}, \hat D_{0|k}= \hat D_{k}, \\
		& 0 < {\gamma}_{k} \le 1,\  0 < {\gamma}_{j,k} \le 1, \, \theta_k \in \Theta, \ u_{i|k} \in {\mathbb{U}},
	\end{align}
\end{subequations}
where $ \ i \in \mathbb{N}_{0:{N}}$, $N$ is the prediction horizon. Regarding self-collision, as mentioned in \cite{mpc_safe3}, we consider this issue by constraining joint positions within a specific set as $\theta_k \in \Theta$, where $\Theta =\{\theta_k \in \mathbb{R}^n \ | \ \theta_{\rm{self,min}} \leq \theta_k \leq \theta_{\rm{self,max}}\}$, $\theta_{\rm{self,min}}$ and $\theta_{\rm{self,max}}$ represent the minimum and maximum joint positions to avoid self-collision, respectively. 

It should be highlighted that $\gamma_{k}$ and $\gamma_{j,k}$ are the optimization variables, which are dynamically optimized throughout the receding horizon optimization process, further enhancing the flexibility and adaptiveness of the proposed CBFSC (\ref{op}d) and (\ref{op}e). The optimal solution to the optimization problem (\ref{op}) is defined as
$\bm{u}^ *_{k}= \begin{bmatrix}
	u^ *_{0|k}, &\!\!\! \!u^ *_{1|k}, &\!\!\!\cdots, & \!\!\! \!u^ *_{ N|k}
\end{bmatrix},$
%\begin{equation}\label{eq1}
%	\bm{u}^ *_{k}= 
%	\begin{bmatrix}
%		u^ *_{0|k}, &\!\!\! \!u^ *_{1|k}, &\!\!\!\cdots, & \!\!\! \!u^ *_{ N|k}
%	\end{bmatrix},
%\end{equation}
the first action, $u^*_k= u^ *_{0|k}$, denotes the desired joint velocities for the joint controller of the robot manipulator.
%At the next time step $k+1$, update the initial conditions (\ref{op}b), (\ref{op}c) and (\ref{op}f), and then repeat solving the optimization problem (\ref{op}) to obtain the optimal control input. 
The framework of the proposed FASM control methodology is illustrated in Fig. 2. 
%the specific design of each module will be presented step by step.
\begin{figure}[t]
	\centering{\includegraphics[width=0.45\textwidth]{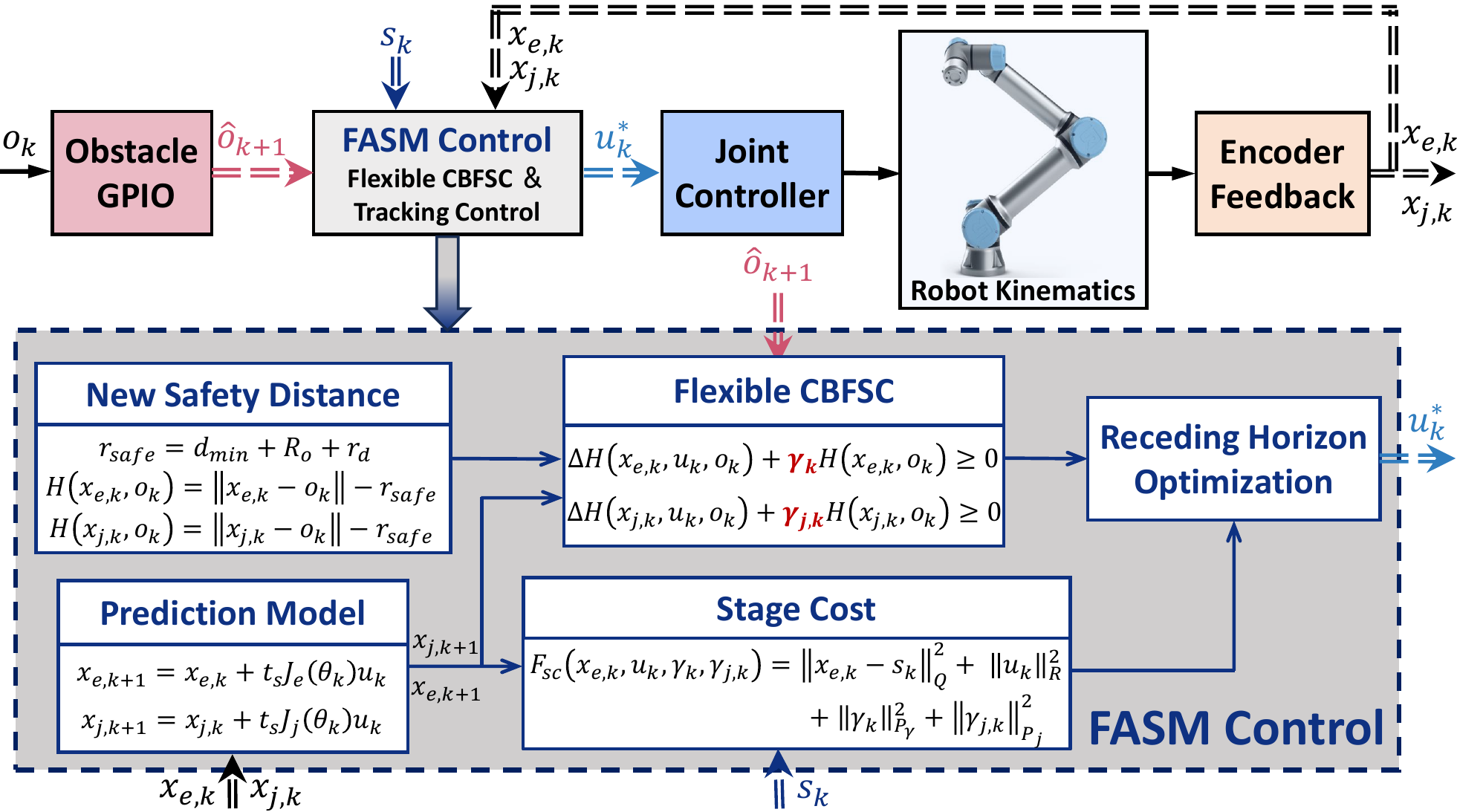}}
	\caption{Framework of the proposed FASM control.}
	\label{fig}
\end{figure}
\section{Experimental Studies}
In this section, to illustrate the effectiveness of the proposed FASM control, experiments are conducted in various scenarios using a UR5 manipulator. Detailed experimental results can also be found in our video on the website. \footnote{[Online]. Available: \url{https://youtu.be/kNY5oEyxj-4}.}
\vspace{-3mm}
\subsection{Experimental Setup and Controller Settings}
The experimental setup includes a UR5 manipulator, a mobile carrier, a motor and two obstacles, as depicted in Fig. 3. Specifically, a cuboid measuring 22 cm in length, 32 cm in height, and 6 cm in width represents the large obstacle, while a cube with edges of 10 cm represents the small obstacle. The software framework utilizes Robot Operating System 2 (ROS2) and MoveIt 2. The execution was performed on a host PC equipped with an i7-10750H CPU and 32GB of RAM.

We use the basic link of the manipulator as the origin of the world coordinate system. We select critical points on each link of the robot manipulator, totaling six critical points on the whole body of the robot manipulator. Table I presents the initial (point A) and desired poses (point B) of the end effector under scenarios involving small and large obstacles. The initial position of the large obstacle relative to the world coordinate is $\begin{bmatrix} 0.60 & -0.40 & 0.18 \end{bmatrix}^{\rm T}$m, while the small obstacle is positioned at $\begin{bmatrix} 0.60 & -0.40 & 0.16 \end{bmatrix}^{\rm T}$m. To comprehensively evaluate dynamic obstacle scenarios, we designed a motor-driven mechanism capable of regulating obstacle movement speeds on the Y-axis at two distinct levels: 6.5 cm/s (slow moving) and 14.5 cm/s (fast moving).
\vspace{-2mm}
\begin{table}[h]
	\renewcommand{\arraystretch}{1}
	\caption{Initial and Desired Poses of the End-Effector}
	\centering
	\label{table_1}
	\resizebox{\columnwidth}{!}{
		\begin{tabular}{ccccccccc}
			\midrule
			%\multicolumn{8}{c}{Cartesian poses of the end-effector} \\ \midrule
			&  &$x$ (m) &$y$ (m)  &$z$ (m) &$q_\omega$  &$q_x$  &$q_y$  &$q_z$  \\ \midrule
			&Point A (Small obstacle) &0.620  &0.368  &0.170  &0.6902 &-0.1499  &0.6914 &0.1519\\ \midrule
			&Point A (Large obstacle) &0.620  &0.368  &0.190  &0.6902 &-0.1499  &0.6914 &0.1519\\ \midrule
			&Point B (Small obstacle)  &0.580  &-0.351  &0.200  &0.6902 &-0.1499  &0.6914 &0.1519  \\ \midrule
			&Point B (Large obstacle)  &0.520  &-0.351  &0.220  &0.6902 &-0.1499  &0.6914 &0.1519  \\ \midrule
		\end{tabular}
	}
\end{table}

The number of degrees of freedom $n=6$. The sampling period $t_s=0.04$ s. In the experiments a three-order GPIO is adopted, the observer gains are designed as $\alpha_1=5$, $\alpha_2=10$, and $\alpha_3=2$. Parameters $\eta=0.9999$, $c_1=3.23$, $c_2=91.68$, $\phi_0=5.33$, $\delta=0.8$, $d_{\rm min}=0.1$ cm. For the small obstacle, $R_o=8.66$ cm. For the large obstacle, $R_o=19.65$ cm. The controller parameters are assigned as $Q=2000$, $R=50$. The settings for $N$ and $P_\gamma$ can be found in each experimental scenario, the initial decay rate is $\gamma_{0}=\gamma_{j,0}=0.001$. The constraints on joint velocities are represented as $\mathbb{U}=\{u \in \mathbb{R}^6 \ | -0.6I_{6 \times 1} \leq u \leq 0.6I_{6 \times 1}\}$. To prevent self-collision, we restrict the position of the fourth joint to be between -2 and 2 rad, the position of the fifth joint to be between -2 and 0 rad, and the remaining joints to be between -2.5 and 2.5 rad. 
\begin{figure}[t]
	\centering{\includegraphics[width=0.45\textwidth]{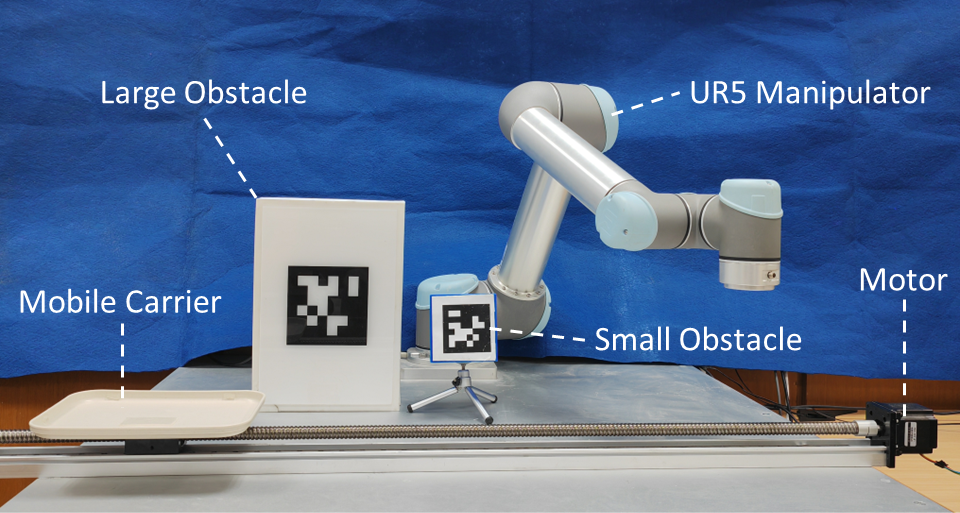}}
	\caption{The experimental setup.}
	\label{fig}
\end{figure}
\vspace{-2mm}
\subsection{Experiment 1: Comparisons Between FASM Control and MPC Method}
In this subsection, we compare the proposed FASM control with the classic MPC method \cite{mpccbf1} in scenarios involving small and large obstacles with rapid movements, respectively. The safety criteria in the MPC method are designed as \( H(x_{e,k}, o_{k}) \geq 0 \), the controller parameters in MPC being the same as those in the proposed FASM control.

As shown in Fig. 4, the proposed FASM control method successfully avoids the small and large obstacles that move quickly, even with the prediction horizon $N=1$. In contrast,  the MPC method fails to achieve dynamic obstacle avoidance in scenarios involving small and large obstacles under $N=1$. This failure can be attributed to the inherent limitations of the MPC in triggering obstacle avoidance action within a shorter prediction horizon. The safety criteria in MPC become active only when the robot manipulator is in close proximity to the obstacle. In scenarios with fast-moving dynamic obstacles, there might not be adequate time for the robot manipulator to avoid obstacles when they are extremely close. In our tests, expanding the prediction horizon to $N=7$, the MPC method effectively avoids the dynamic obstacle. (This result is demonstrated in the video.) This highlights the effectiveness of increasing the prediction horizon in enhancing obstacle avoidance capabilities but results in increased computational requirements.
%Expanding the prediction horizon to $N=7$, the BMPC effectively avoids the dynamic obstacle, exhibiting performance comparable to FAST control under $P_\gamma=150$, $N=1$. This highlights the effectiveness of increasing the prediction horizon to $N=7$ in enhancing obstacle avoidance capabilities, but results in increased computational requirements.
%the first column of the images shows the curve of position error changing with time, the second column shows the curve of posture error changing with time, and the third column shows the curve of joint velocity changing with time. 
Moreover, as illustrated in Fig. 5, in both scenarios, the position and posture errors of the proposed FASM control eventually converge to zero, and the joint velocities satisfy the constraints. This indicates the successful completion of the reference tracking and constraint satisfaction tasks.
\begin{figure*}[t]
	\centering
	\setcounter{subfigure}{0}	
	\subfigure{
		\begin{minipage}[t]{1\linewidth}
			\centering
			\includegraphics[width=6.6in]{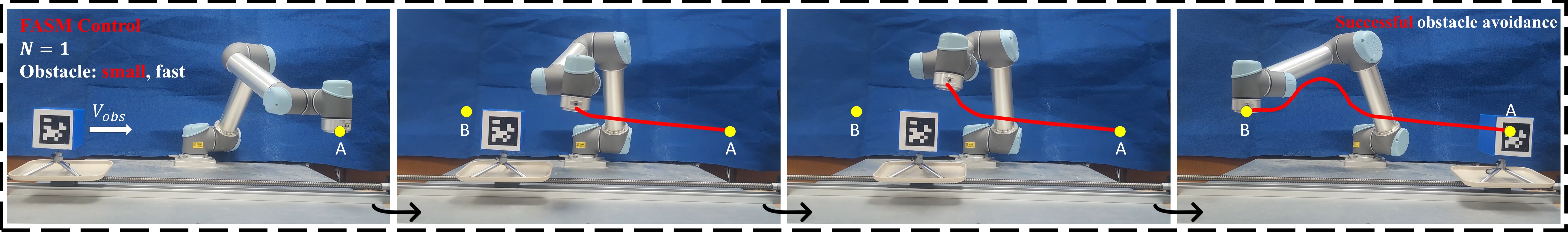}
			%\caption{fig1}
		\end{minipage}%
	}%	\setcounter{subfigure}{0}
	%µÚËÄÐÐÍ¼Æ¬Õ¹Ê¾
	\vspace{-2.1mm}
	\setcounter{subfigure}{0}	
	\subfigure{
		\begin{minipage}[t]{1\linewidth}
			\centering
			\includegraphics[width=6.6in]{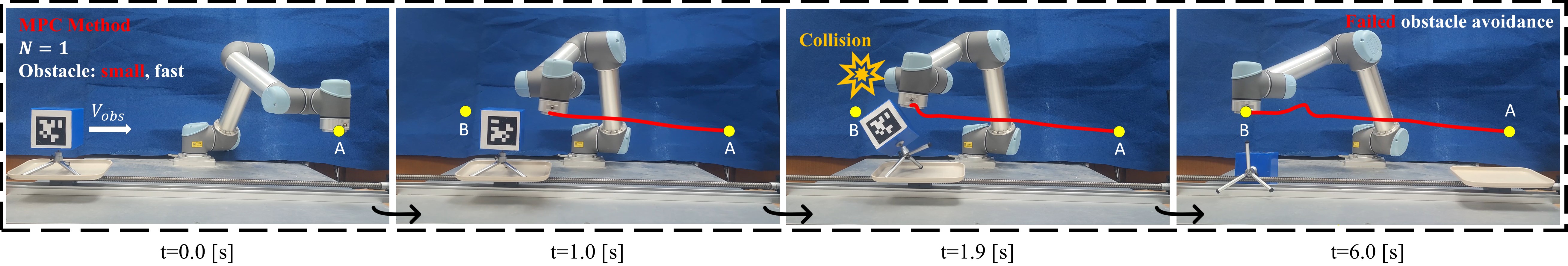}
			%\caption{fig1}
		\end{minipage}%
	}%	\setcounter{subfigure}{0}
\setcounter{subfigure}{0}	
\subfigure{
	\begin{minipage}[t]{1\linewidth}
		\centering
		\includegraphics[width=6.6in]{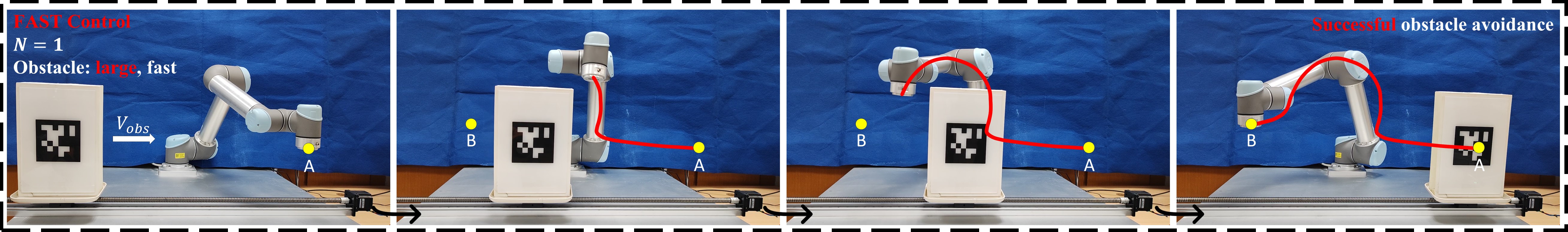}
		%\caption{fig1}
	\end{minipage}%
}%	\setcounter{subfigure}{0}
\vspace{-2.1mm}
\setcounter{subfigure}{0}	
\subfigure{
	\begin{minipage}[t]{1\linewidth}
		\centering
		\includegraphics[width=6.6in]{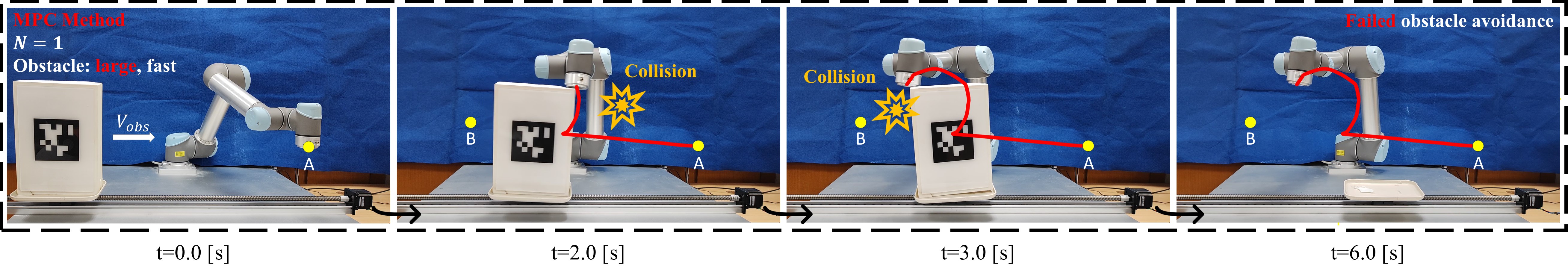}
		%\caption{fig1}
	\end{minipage}%
}%	\setcounter{subfigure}{0}
	\caption{Experiment 1: Frame-by-frame plots in the scenarios of fast-moving small and large obstacles (top: small obstacle; bottom: large obstacle). (Only the trajectory from point A to point B is shown.)}
	\label{fig:result_include1}
\end{figure*}
\begin{figure*}[h!]
	\centering
	%µÚÒ»ÐÐÍ¼Æ¬Õ¹Ê¾
	\setcounter{subfigure}{0}	
	% µÚ¶þÐÐÍ¼Æ¬Õ¹Ê¾
	\subfigure{
		\begin{minipage}[t]{0.23\linewidth}
			\centering
			\includegraphics[width=1.8in]{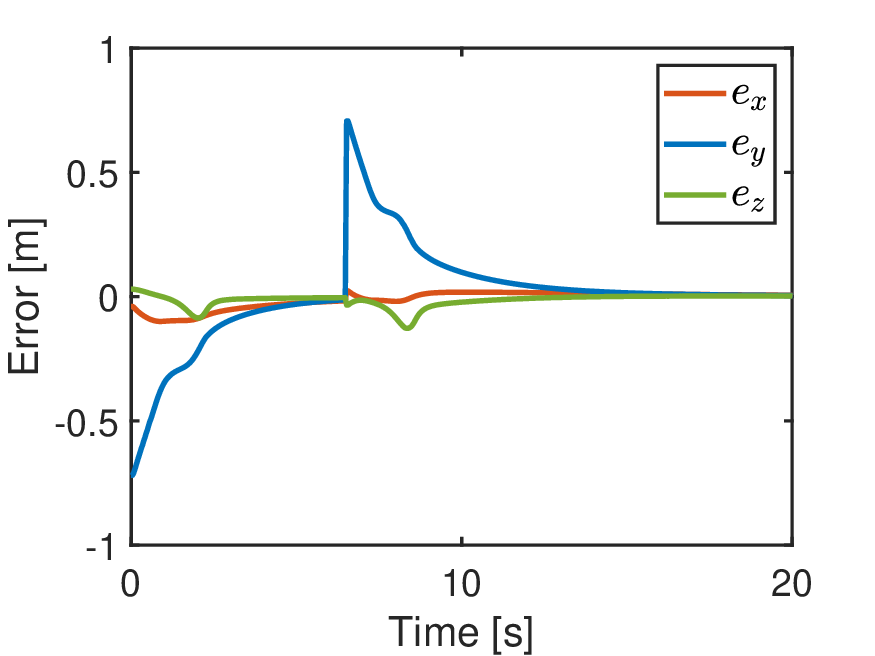}
		\end{minipage}%
	}%
	\subfigure{
		\begin{minipage}[t]{0.23\linewidth}
			\centering
			\includegraphics[width=1.8in]{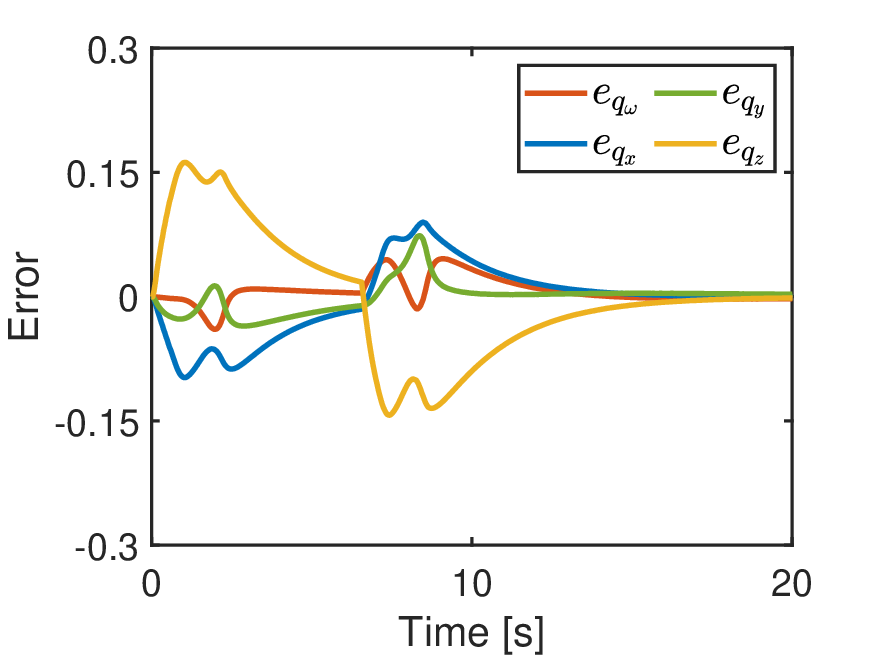}
		\end{minipage}%
	}%
	\subfigure{
		\begin{minipage}[t]{0.23\linewidth}
			\centering
			\includegraphics[width=1.8in]{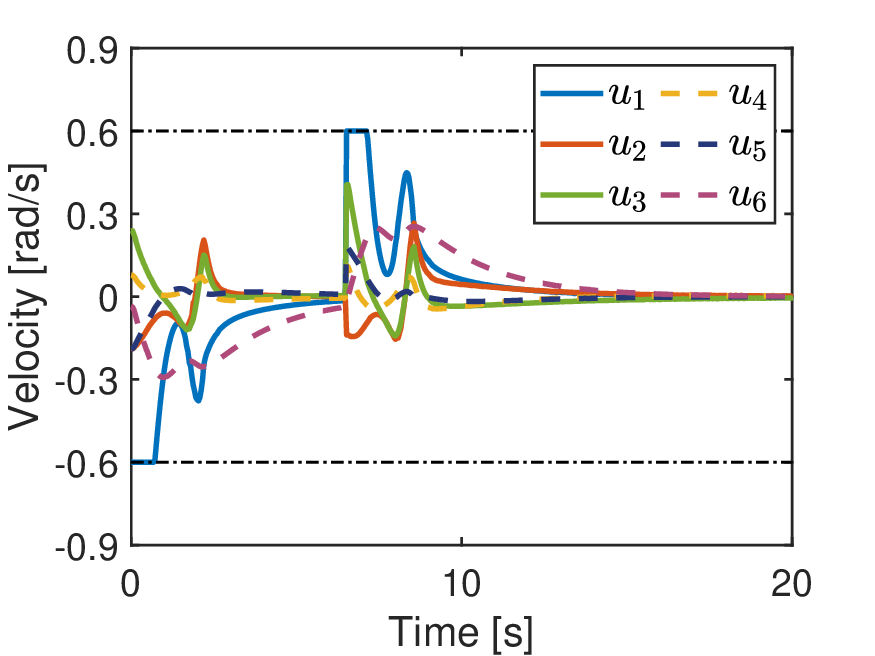}
		\end{minipage}%
	}%
	\subfigure{
		\begin{minipage}[t]{0.25\linewidth}
			\centering
			\includegraphics[width=1.85in]{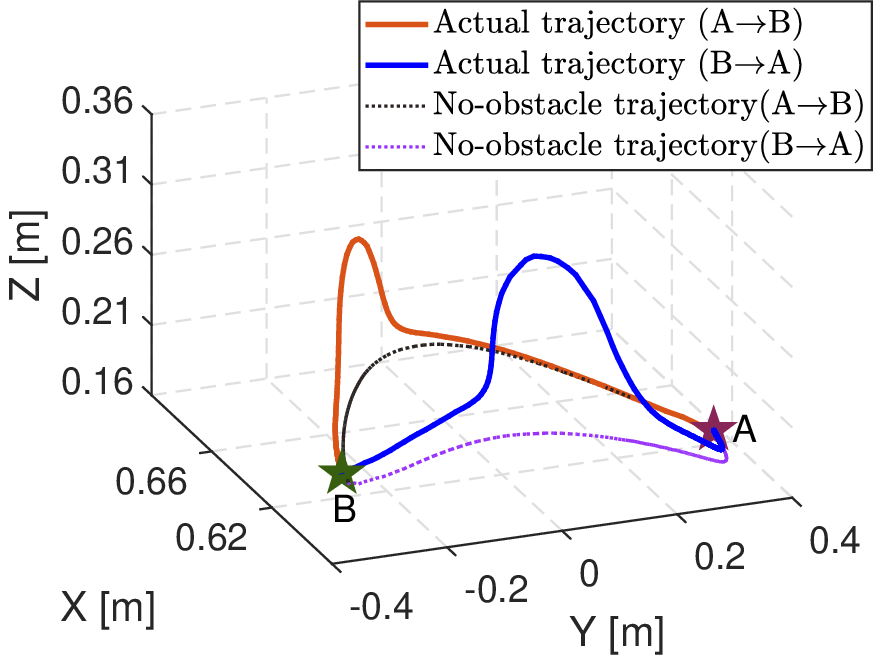}
			%\caption{fig1}
		\end{minipage}%
	}%
	\vspace{-3.5mm}
	\setcounter{subfigure}{0}	
	% µÚÈýÐÐÍ¼Æ¬Õ¹Ê¾
	\subfigure[Position errors]{
		\begin{minipage}[t]{0.23\linewidth}
			\centering
			\includegraphics[width=1.8in]{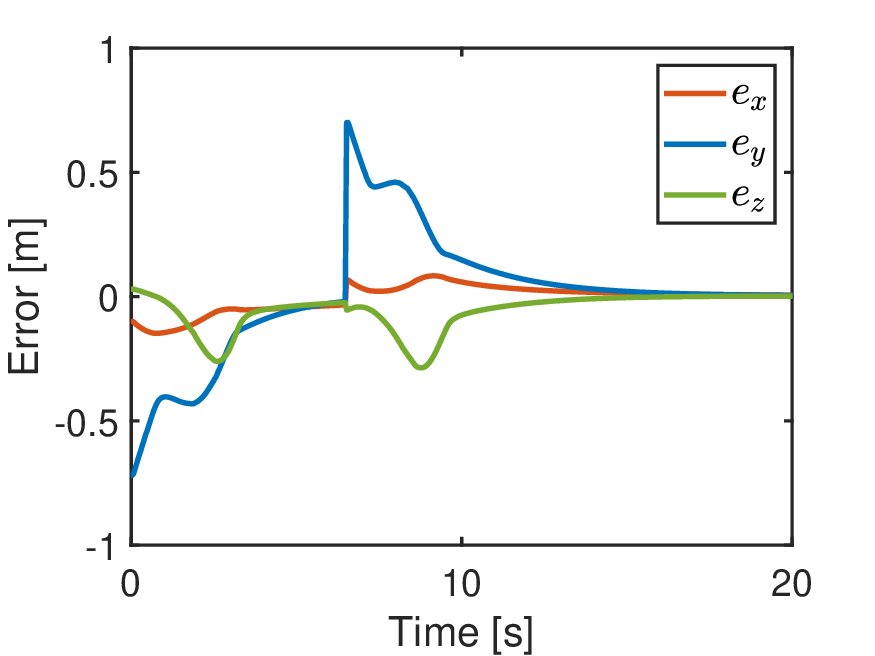}
		\end{minipage}%
	}%
	\subfigure[Quaternion errors]{
		\begin{minipage}[t]{0.23\linewidth}
			\centering
			\includegraphics[width=1.8in]{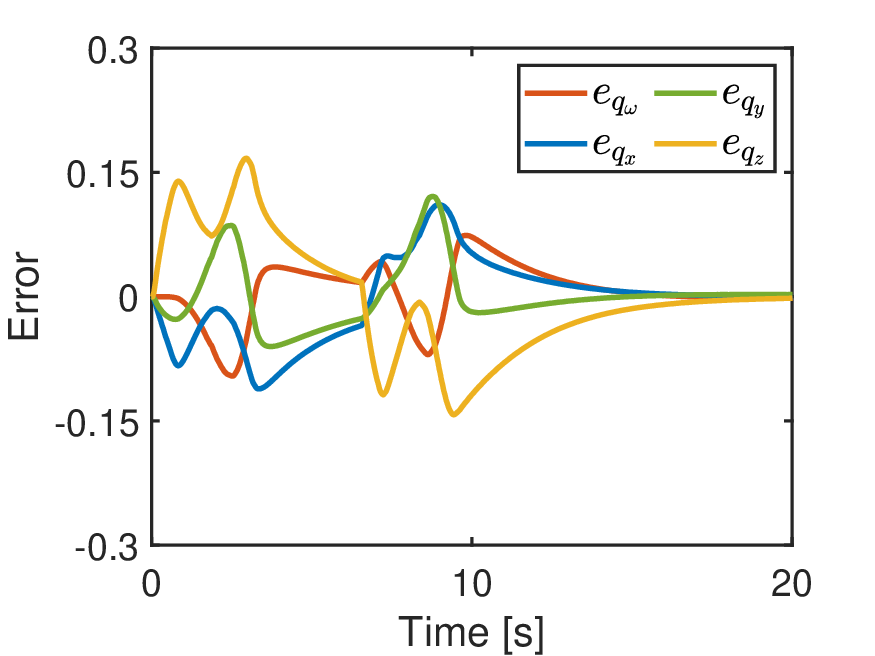}
		\end{minipage}%
	}%
	\subfigure[Joint velocities]{
		\begin{minipage}[t]{0.23\linewidth}
			\centering
			\includegraphics[width=1.8in]{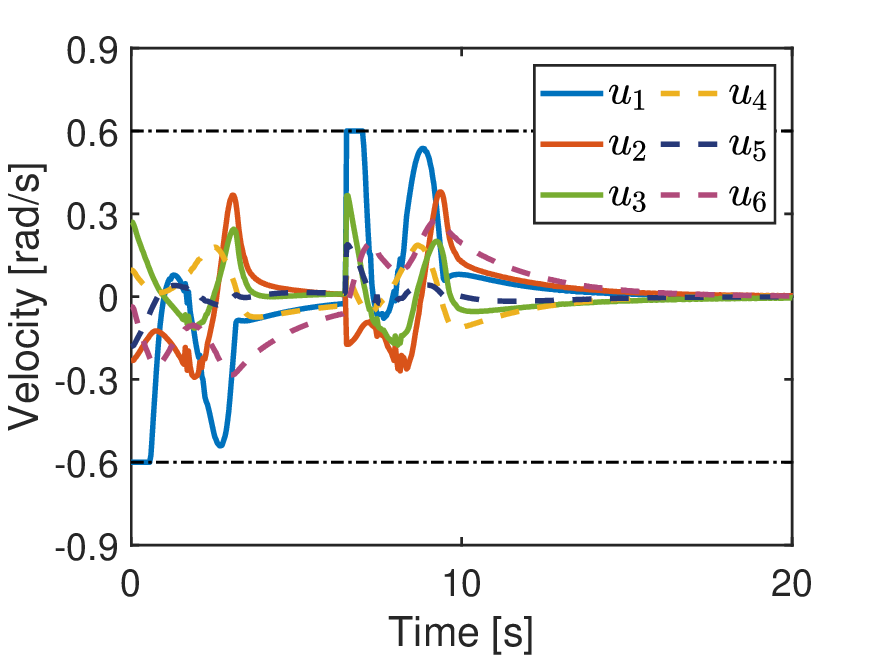}
		\end{minipage}%
	}%
	\subfigure[Trajectories of the end-effector]{
		\begin{minipage}[t]{0.25\linewidth}
			\centering
			\includegraphics[width=1.85in]{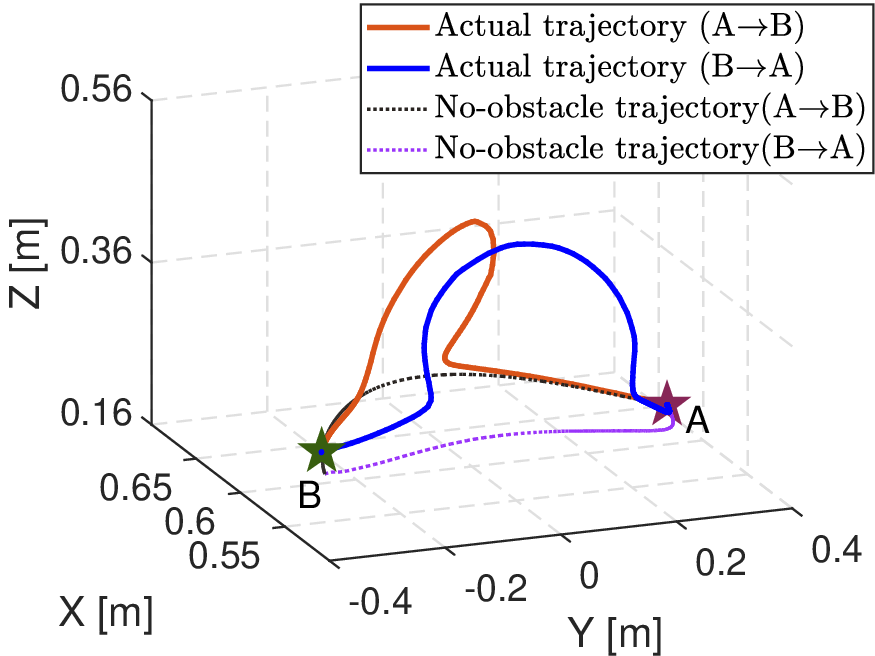}
			%\caption{fig1}
		\end{minipage}%
	}%
	% Ìí¼ÓÌâ×¢£¬¼´¶ÔÕâ¸öÍ¼Æ¬µÄËµÃ÷
	\caption{Experiment 1: Position errors, quaternion errors, joint velocities and end-effector trajectories of the proposed FASM control in the scenarios of fast-moving small and large obstacles (top: small obstacle; bottom: large obstacle).}
	\label{fig:result_include1}
\end{figure*}

In general, compared to the MPC method, the proposed FASM control can employ a smaller prediction horizon to avoid obstacles, offering a more proactive safety strategy. This capability arises from the ability of the proposed flexible CBFSC to trigger the robot to act even when it is positioned far from obstacles by dynamically optimizing $\gamma_{k}$. %Additionally, the proposed method can decrease the computational load as there is no necessity to set a larger prediction horizon.
\vspace{-3mm}
\subsection{Experiment 2: Comparisons Across Different $P_\gamma$}
In this subsection, we investigate the impact of different values of $P_\gamma$ on the performance of the proposed FASM control. We specifically concentrate on the scenario of a small slow-moving obstacle under $P_\gamma=150$, $P_\gamma=1000$, and $P_\gamma=2000$. This scenario also assesses the performance of the proposed FASM control at different obstacle velocities.
\begin{figure*}[t]
	\centering
	\subfigure[The values of $H(x_{e,k}, o_{k})$.]{
		\begin{minipage}[t]{0.45\linewidth}
			\centering
			\includegraphics[width=3.3in]{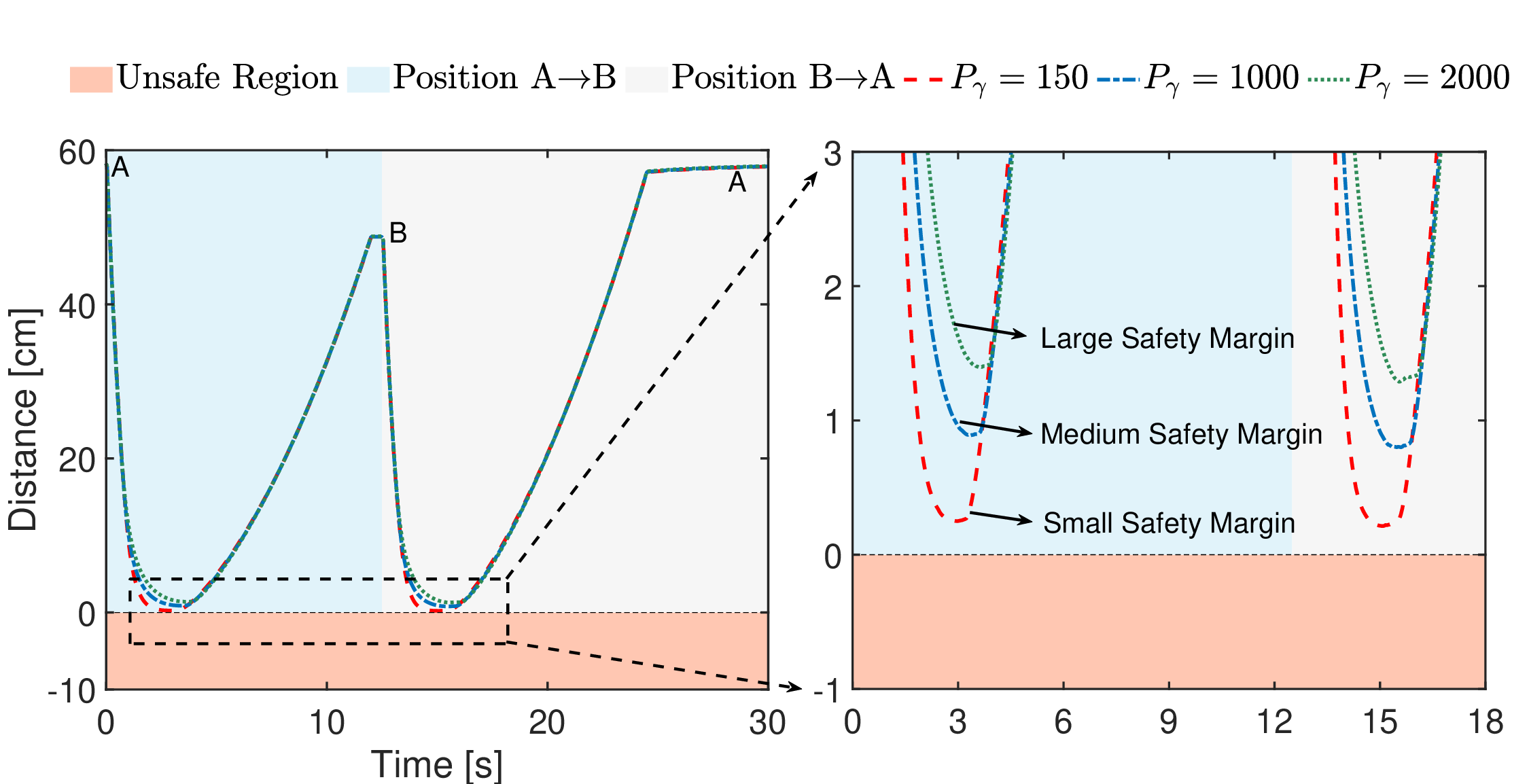}
			%\caption{fig1}
		\end{minipage}%
	}%
	\subfigure[The values of $\gamma_k$.]{
		\begin{minipage}[t]{0.25\linewidth}
			\centering
			\includegraphics[width=1.9in]{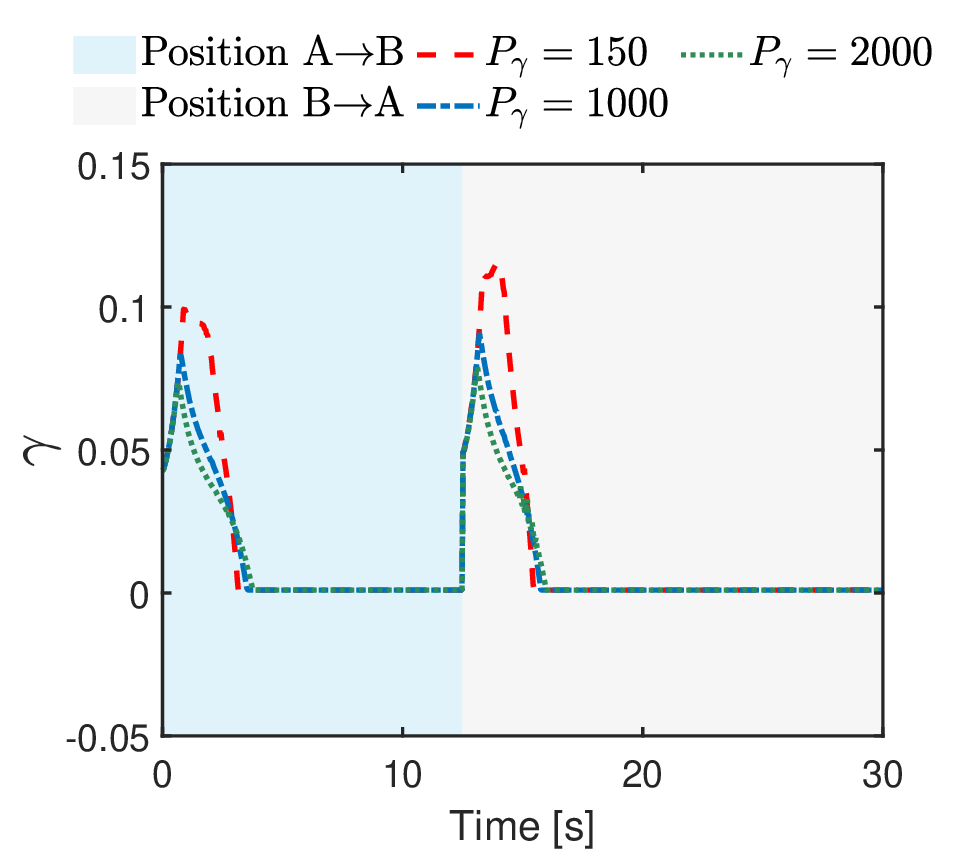}
			%\caption{fig2}
		\end{minipage}%
	}%
	\subfigure[Velocity estimation of the obstacle.]{
		\begin{minipage}[t]{0.25\linewidth}
			\centering
			\includegraphics[width=1.72in]{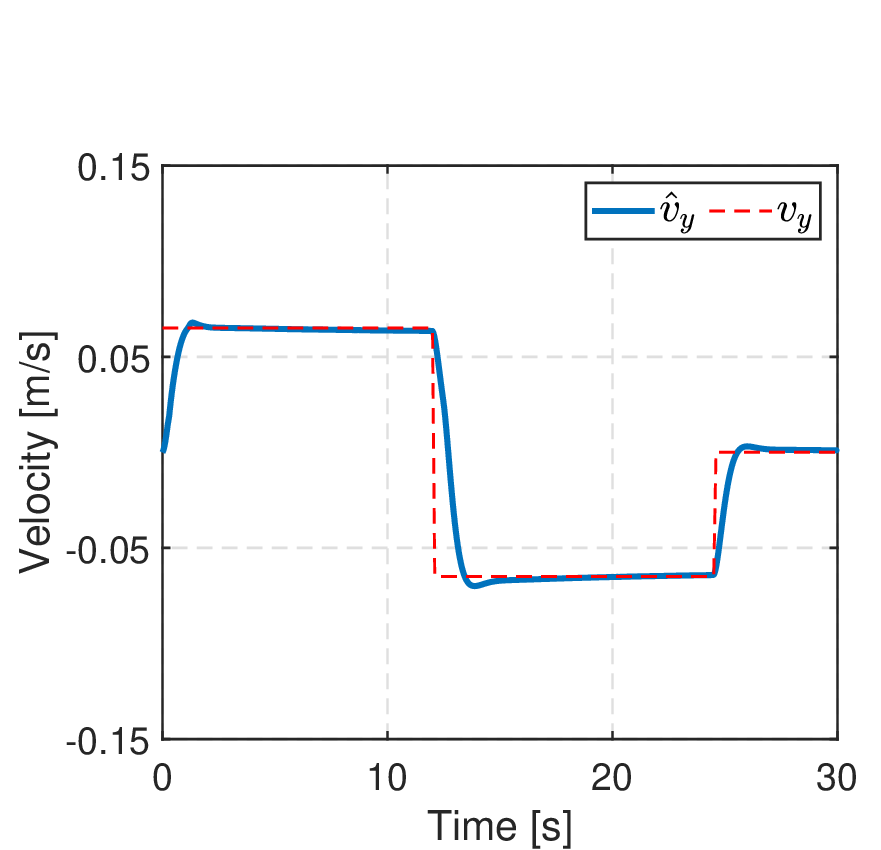}
			%\caption{fig2}
		\end{minipage}%
	}%
	\centering
	\caption{Experiment 2: The experimental results of the FASM control in the scenario of the slow-moving small obstacle under different parameters. (The test path is A$\rightarrow$B$\rightarrow$A.)}
\end{figure*}
\begin{figure*}[t]
	\centering
	%µÚÒ»ÐÐÍ¼Æ¬Õ¹Ê¾
	\subfigure{
		\begin{minipage}[t]{1\linewidth}
			\centering
			\includegraphics[width=6.6in]{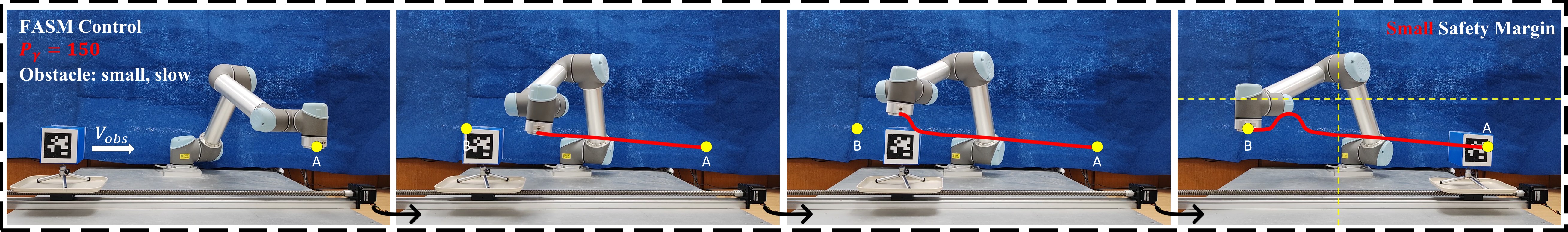}
			%\caption{fig1}
		\end{minipage}%
	}%	\setcounter{subfigure}{0}
	%µÚ¶þÐÐÍ¼Æ¬Õ¹Ê¾
	\vspace{-2.1mm}
	\setcounter{subfigure}{0}	
	\subfigure{
		\begin{minipage}[t]{1\linewidth}
			\centering
			\includegraphics[width=6.6in]{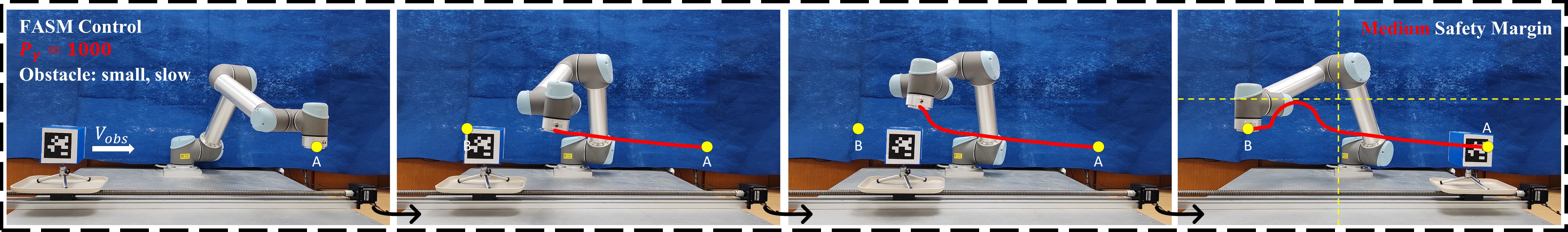}
			%\caption{fig1}
		\end{minipage}%
	}%	\setcounter{subfigure}{0}
	\vspace{-2.1mm}
	\setcounter{subfigure}{0}	
	\subfigure{
		\begin{minipage}[t]{1\linewidth}
			\centering
			\includegraphics[width=6.6in]{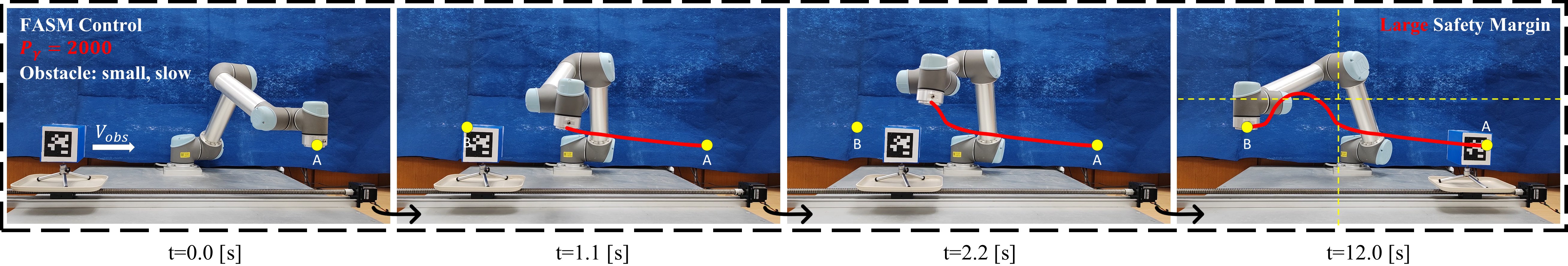}
			%\caption{fig1}
		\end{minipage}%
	}%	\setcounter{subfigure}{0}
	% Ìí¼ÓÌâ×¢£¬¼´¶ÔÕâ¸öÍ¼Æ¬µÄËµÃ÷
	\caption{Experiment 2: Frame-by-frame plots in the scenario of the slow-moving small obstacle under different parameters. (Only the trajectory from point A to point B is shown.)}
	\label{fig:result_include1}
\end{figure*}
\begin{figure*}[t]
	\centering
	%µÚÒ»ÐÐÍ¼Æ¬Õ¹Ê¾
	\subfigure{
		\begin{minipage}[t]{0.23\linewidth}
			\centering
			\includegraphics[width=1.8in]{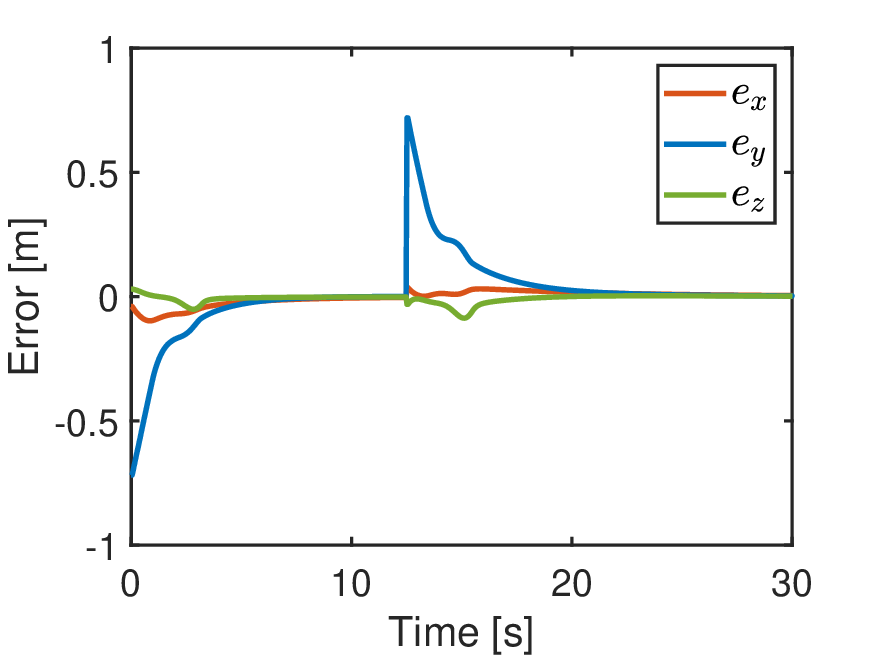}
			%\caption{fig1}
		\end{minipage}%
	}%	\setcounter{subfigure}{0}
	\subfigure{
		\begin{minipage}[t]{0.23\linewidth}
			\centering
			\includegraphics[width=1.8in]{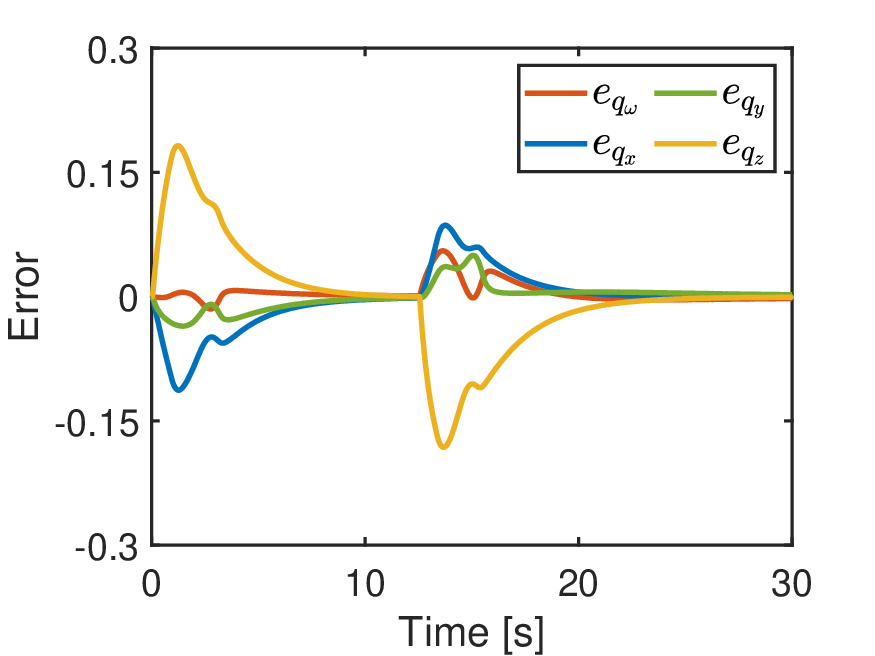}
			%\caption{fig1}
		\end{minipage}%
	}%
	\subfigure{
		\begin{minipage}[t]{0.23\linewidth}
			\centering
			\includegraphics[width=1.8in]{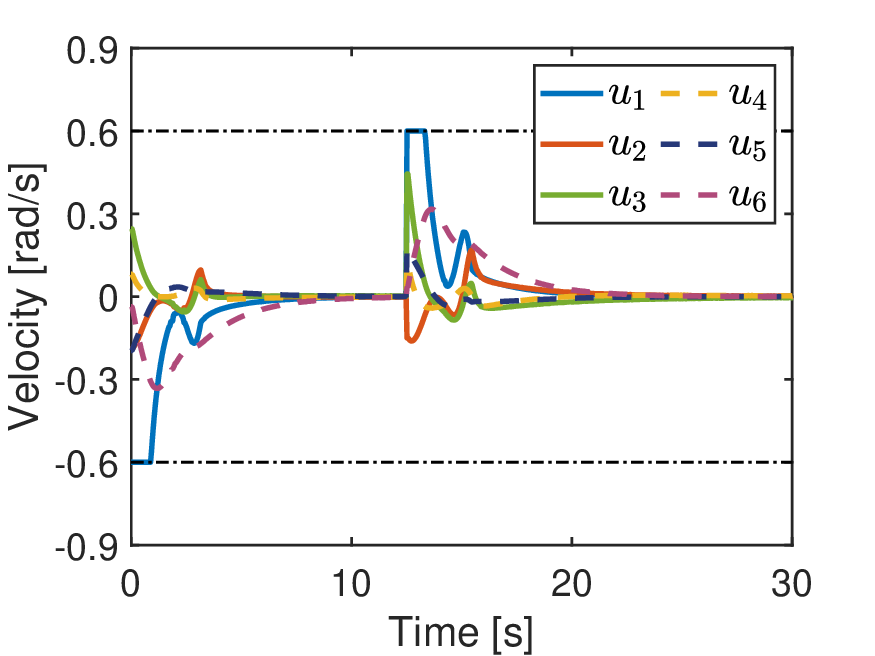}
			%\caption{fig1}
		\end{minipage}%
	}%
	\subfigure{
		\begin{minipage}[t]{0.25\linewidth}
			\centering
			\includegraphics[width=1.85in]{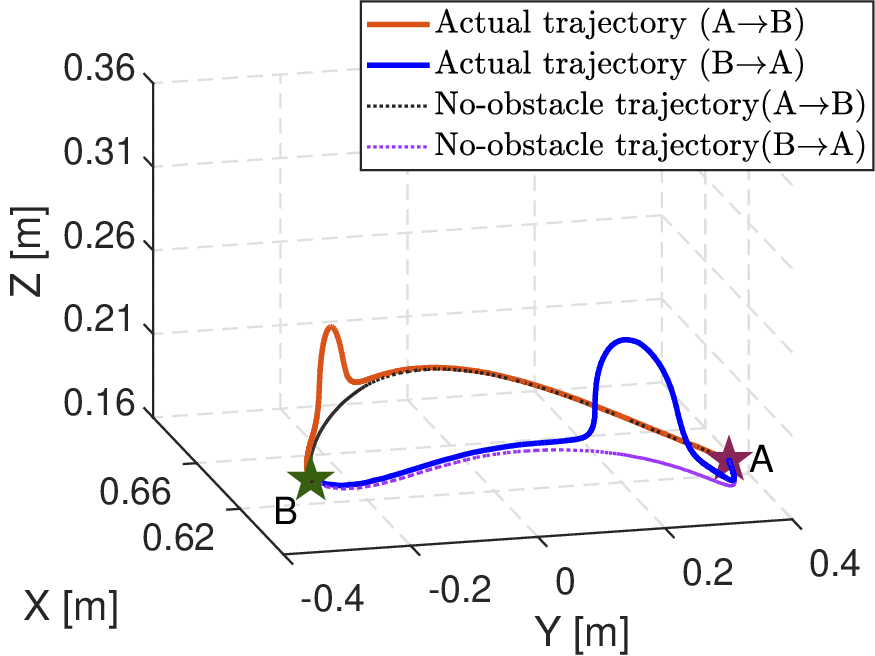}
			%\caption{fig1}
		\end{minipage}%
	}%
	 %µÚ¶þÐÐÍ¼Æ¬Õ¹Ê¾
		\vspace{-3.5mm}
		\setcounter{subfigure}{0}	
		\subfigure{
			\begin{minipage}[t]{0.23\linewidth}
					\centering
					\includegraphics[width=1.8in]{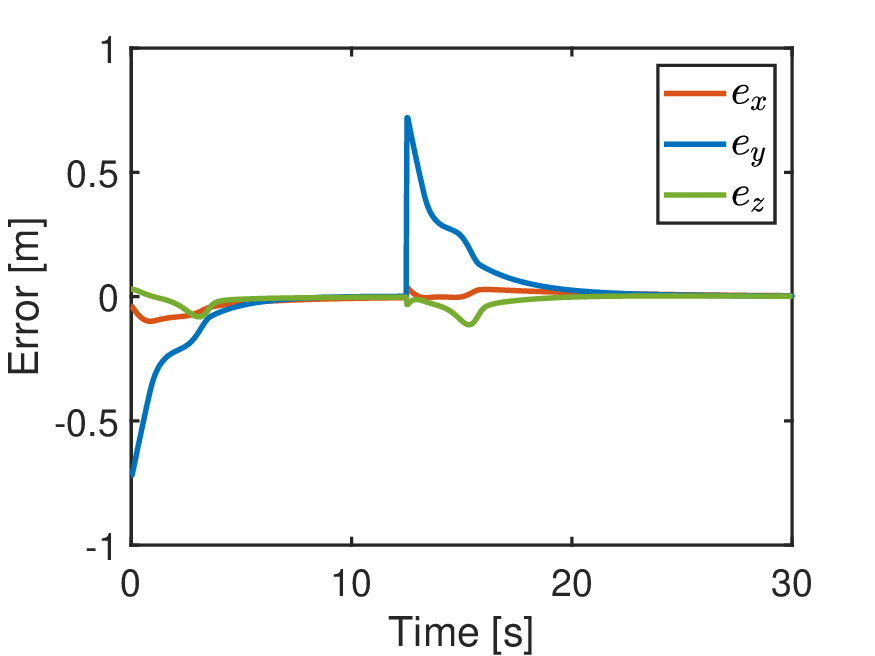}
						%\caption{fig1}
				\end{minipage}%
			}%	\setcounter{subfigure}{0}
		\subfigure{
				\begin{minipage}[t]{0.23\linewidth}
						\centering
						\includegraphics[width=1.8in]{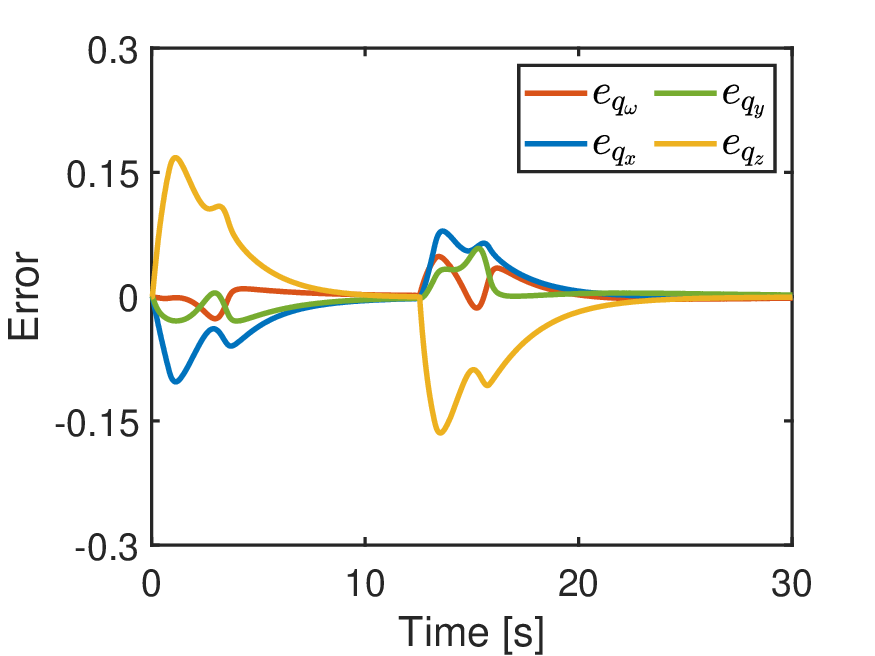}
						%\caption{fig1}
					\end{minipage}%
			}%
		\subfigure{
				\begin{minipage}[t]{0.23\linewidth}
						\centering
						\includegraphics[width=1.8in]{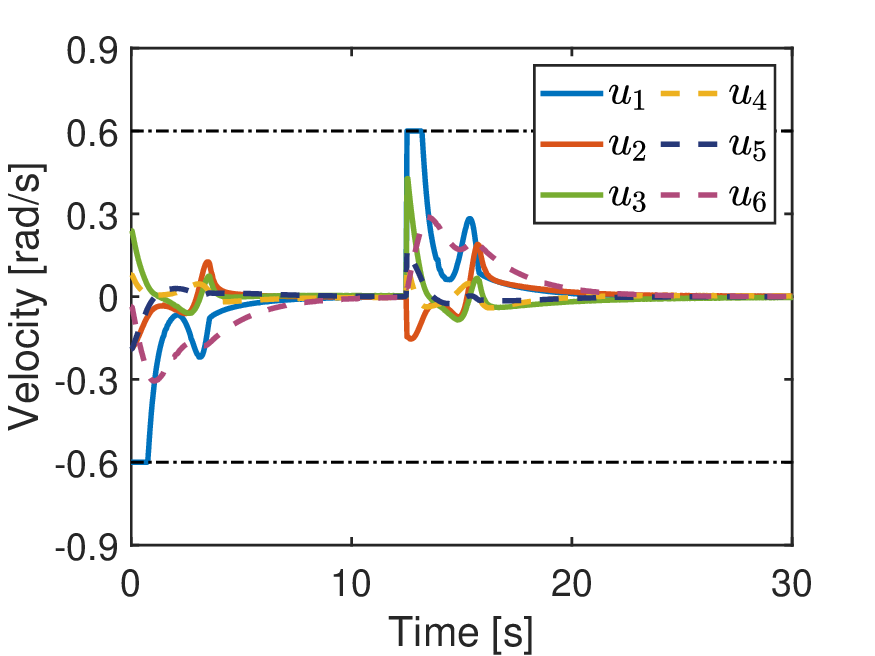}
						%\caption{fig1}
					\end{minipage}%
			}%
		\subfigure{
				\begin{minipage}[t]{0.25\linewidth}
						\centering
						\includegraphics[width=1.85in]{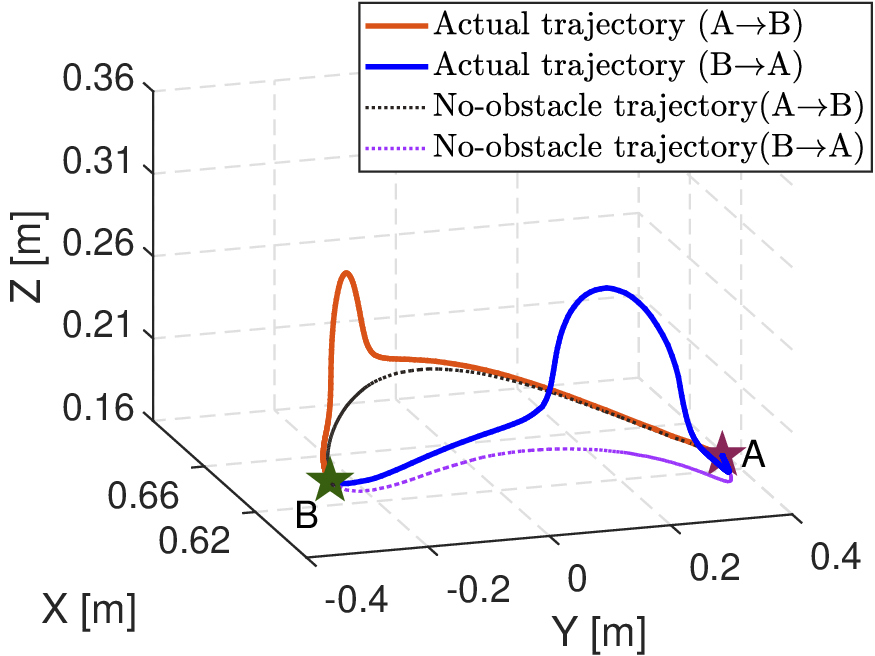}
						%\caption{fig1}
					\end{minipage}%
			}%
	% µÚÈýÐÐÍ¼Æ¬Õ¹Ê¾
	\vspace{-3.5mm}
	\setcounter{subfigure}{0}	
	\subfigure[Position errors]{
		\begin{minipage}[t]{0.23\linewidth}
			\centering
			\includegraphics[width=1.8in]{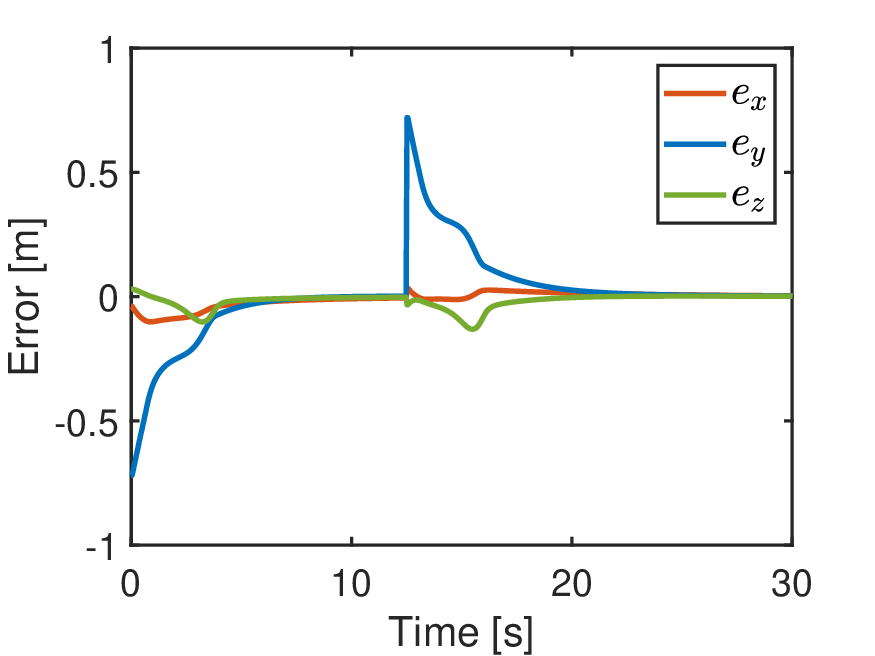}
		\end{minipage}%
	}%
	\subfigure[Quaternion errors]{
		\begin{minipage}[t]{0.23\linewidth}
			\centering
			\includegraphics[width=1.8in]{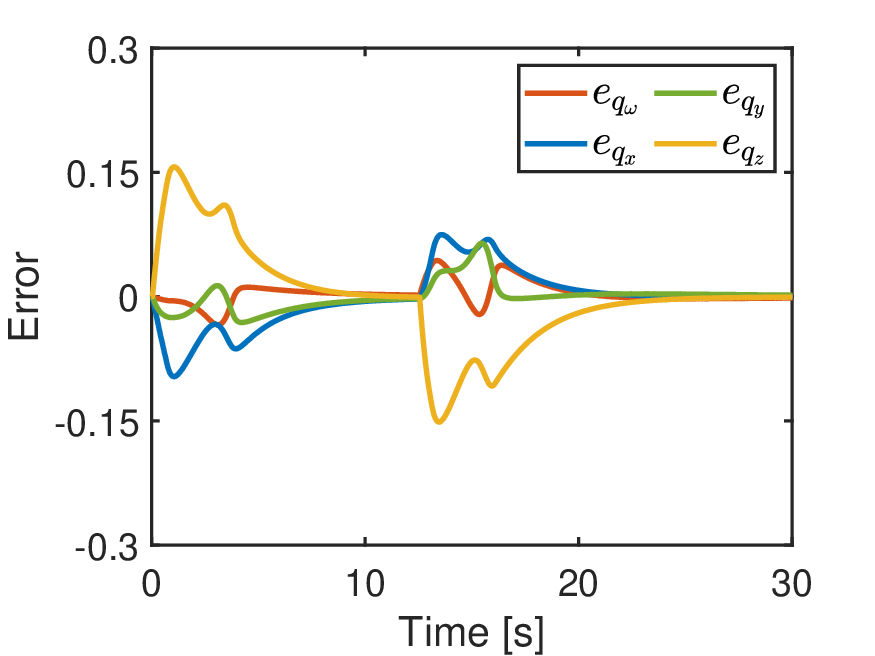}
		\end{minipage}%
	}%
	\subfigure[Joint velocities]{
		\begin{minipage}[t]{0.23\linewidth}
			\centering
			\includegraphics[width=1.8in]{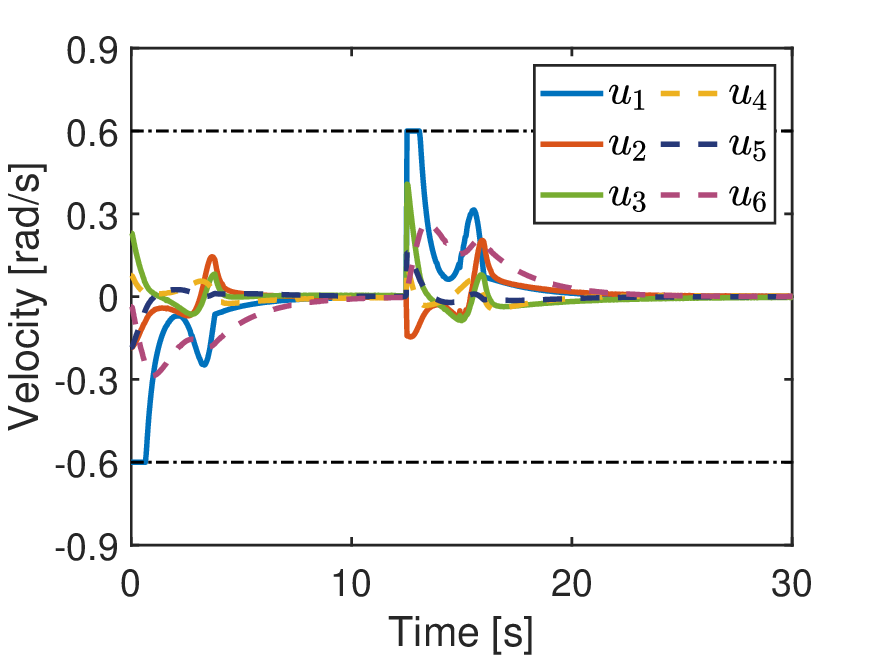}
		\end{minipage}%
	}%
	\subfigure[Trajectories of the end-effector]{
		\begin{minipage}[t]{0.25\linewidth}
			\centering
			\includegraphics[width=1.85in]{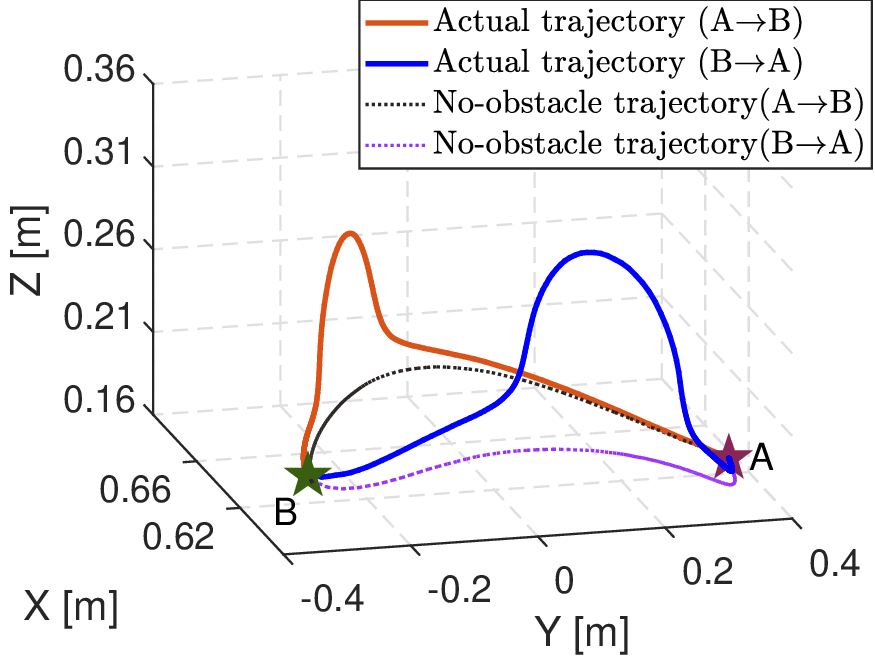}
			%\caption{fig1}
		\end{minipage}%
	}%
	% Ìí¼ÓÌâ×¢£¬¼´¶ÔÕâ¸öÍ¼Æ¬µÄËµÃ÷
	\caption{Experiment 2: Position errors, quaternion errors, joint velocities and end-effector trajectories of the proposed FASM control in the scenario of the slow-moving small obstacle under different values of $P_\gamma$. The first, second, and third lines correspond to $P_\gamma=150$, $P_\gamma=1000$, and $P_\gamma=2000$, respectively.}
	\label{fig:result_include1}
\end{figure*}
\begin{figure*}[t]
	\centering
	\subfigure[The values of $H(x_{e,k}, o_{k})$.]{
		\begin{minipage}[t]{0.45\linewidth}
			\centering
			\includegraphics[width=3.3in]{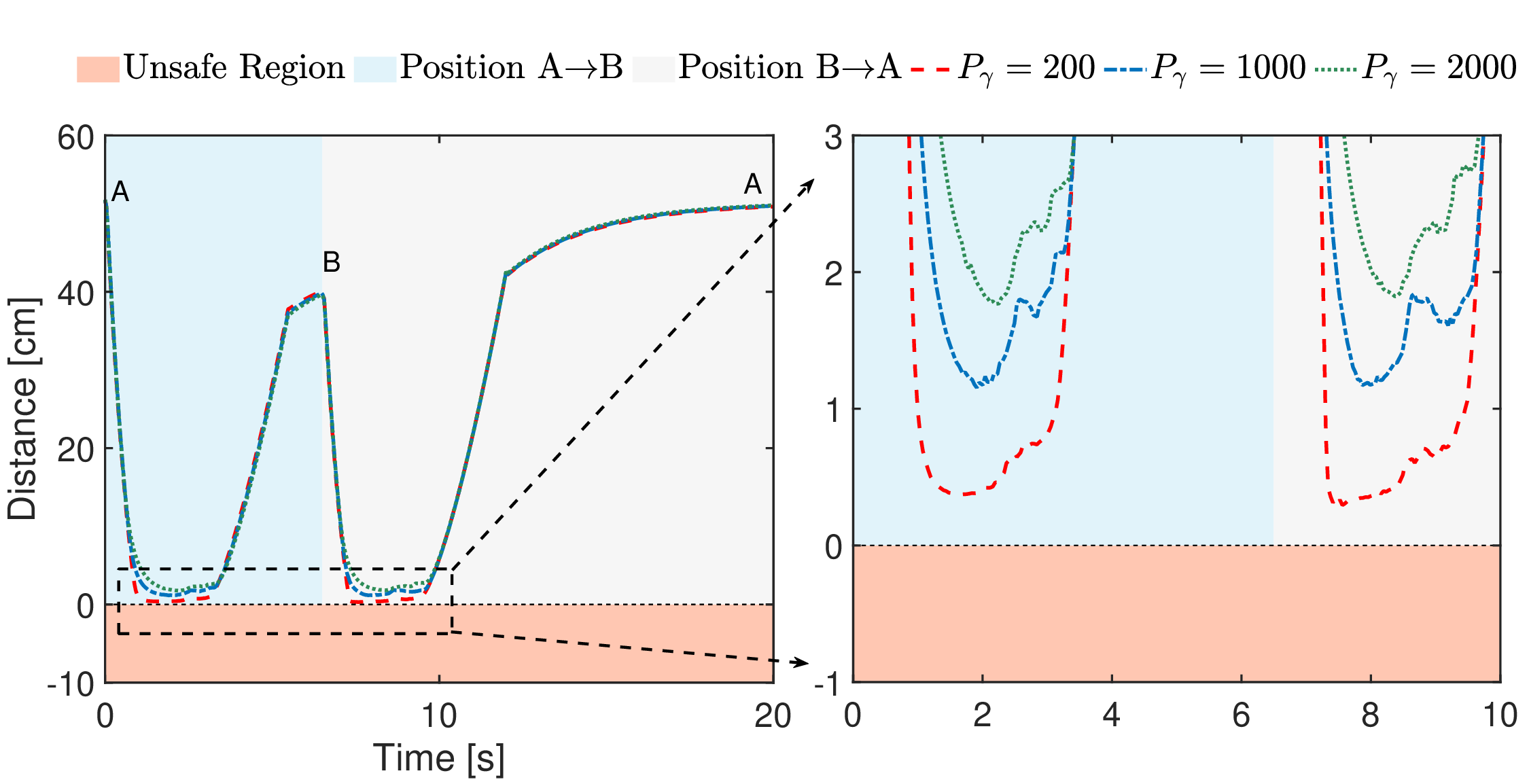}
			%\caption{fig1}
		\end{minipage}%
	}%
	\subfigure[The values of $\gamma_k$.]{
		\begin{minipage}[t]{0.25\linewidth}
			\centering
			\includegraphics[width=1.76in]{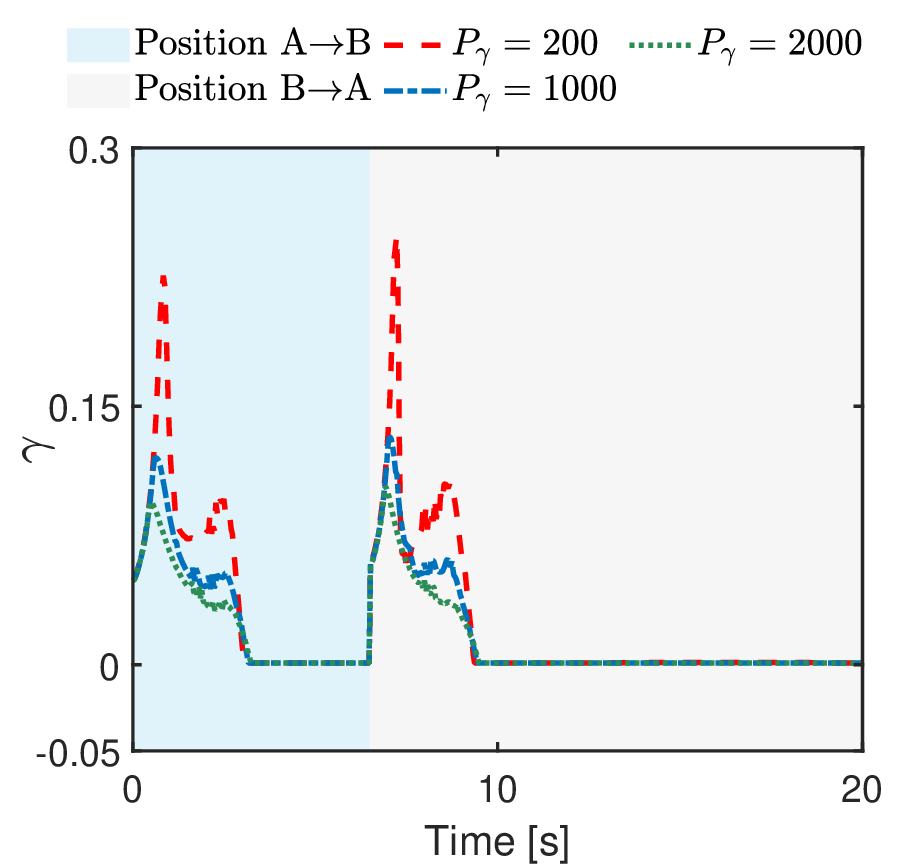}
			%\caption{fig2}
		\end{minipage}%
	}%
	\subfigure[Velocity estimation of the obstacle.]{
		\begin{minipage}[t]{0.25\linewidth}
			\centering
			\includegraphics[width=1.74in]{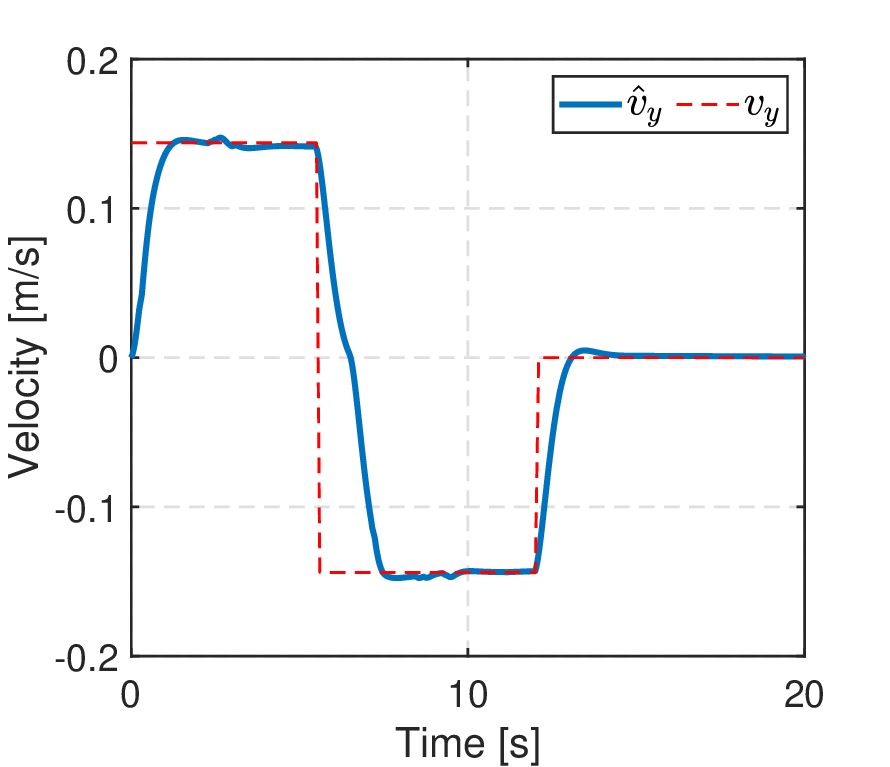}
			%\caption{fig2}
		\end{minipage}%
	}%
	\centering
	\caption{Experiment 3: The experimental results of the FASM control in the scenario of the fast-moving large obstacle under different parameters. (The test path is A$\rightarrow$B$\rightarrow$A.)}
\end{figure*}
%	\vspace{-3.5mm}

As illustrated in Fig. 6(c), the designed GPIO can effectively estimate the velocity of the slow-moving obstacle. As depicted in Figs. 6(a) and 7, different values of $P_\gamma$ provide adjustable safety margins, with larger values corresponding to greater safety margins. Specifically, $P_\gamma=150$, $P_\gamma=1000$, and $P_\gamma=2000$ correspond to small, medium, and large safety margins, respectively. Furthermore, Fig. 6(b) demonstrates that higher $P_\gamma$ values lead to smaller overall values of $\gamma_k$, making the safety constraints more critical and indicating an increased emphasis on safety. This observation can be supported by Fig. 7 and Table II, where a larger $P_\gamma$ results in more advanced avoidance actions and a higher lift of the end-effector. Moreover, in Fig. 6, as the distance between the robot and the obstacle increases, the value of $\gamma_k$ decreases, allowing for earlier triggering of obstacle avoidance actions and thus achieving active safety strategies. Conversely, as the robot approaches the obstacle, the value of $\gamma_k$ is optimized to be larger, inherently relaxing the safety constraints to enhance the feasibility of the optimization problem. In Fig. 8, the position and posture errors of the proposed method eventually converge to zero in each case. Simultaneously, the joint velocities strictly satisfy the constraints, indicating the successful accomplishment of all the tasks.

In general, the proposed method allows for the adjustment of $P_\gamma$ to select various safety margins and dynamically optimizes $\gamma_k$, thereby providing flexible safety constraints. This flexibility is particularly significant as it expands the feasible region, especially when the robot approaches the obstacles.
\vspace{-3mm}
\begin{table}[h]
	\renewcommand{\arraystretch}{0.8}
	\caption{The Performance Under Different Values of $P_\gamma$}
	\centering
	\label{table_1}
	\resizebox{\columnwidth}{!}{
		\begin{tabular}{ccccc}
			\midrule
			&  &$P_\gamma=150$ &$P_\gamma=1000$  &$P_\gamma=2000$\\ \midrule
			&The highest altitude &0.2509 [m] &0.2816 [m]
			&0.3019 [m]\\ \midrule
			&The trigger moment &1.84 [s]&1.56 [s]  &1.16 [s] \\ \midrule
			&Maximum value of $\gamma_{k}$  &0.1151&0.0909  &0.0791 \\ \midrule
		\end{tabular}
	}
\end{table}
\vspace{-3mm}
\subsection{Experiment 3: Performance of Avoiding Large Obstacle}
In this test, we will assess the performance of the proposed FASM control in avoiding the fast-moving large obstacle.

Avoiding fast-moving large obstacles poses a significant challenge, particularly requiring the robot manipulator to initiate avoidance actions at an early stage and ensuring that joint velocities strictly satisfy the constraints. As illustrated in Fig. 9(c), the designed GPIO can effectively estimate the velocity of the fast-moving obstacle. In Figs. 9-11, the proposed FASM control effectively avoids the fast-moving large obstacle across different values of $P_\gamma$, each corresponding to distinct safety margins. As shown by Fig. 10, the proposed approach guides the manipulator to exhibit varying degrees of proactive action as the end-effector approaches the obstacle. In Fig. 9(b), $\gamma_k$ is dynamically optimized to provide flexibility throughout the obstacle avoidance process. As shown by Fig. 11, the position and posture errors of the proposed method eventually converge to zero in each case. Simultaneously, the joint velocities strictly satisfy the constraints, indicating the successful accomplishment of the reference tracking and constraint satisfaction tasks.
\begin{figure*}[t!]
	\centering
	%µÚÒ»ÐÐÍ¼Æ¬Õ¹Ê¾
	\subfigure{
		\begin{minipage}[t]{1\linewidth}
			\centering
			\includegraphics[width=6.6in]{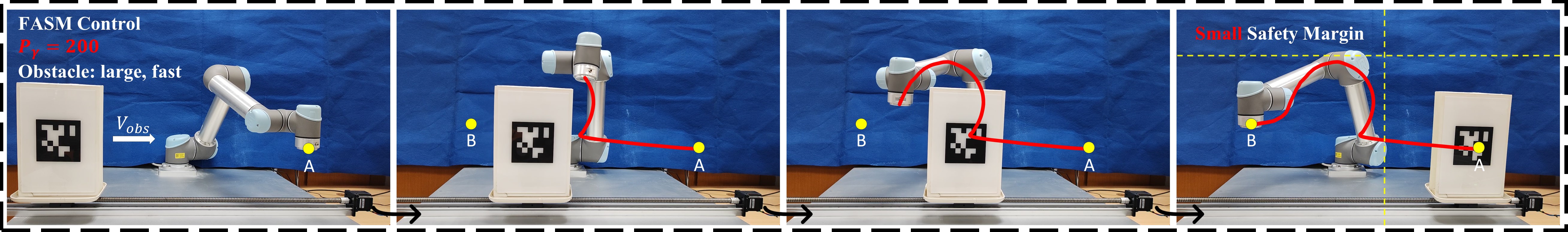}
			%\caption{fig1}
		\end{minipage}%
	}%	\setcounter{subfigure}{0}
	\vspace{-2.1mm}
	\setcounter{subfigure}{0}	
	\subfigure{
		\begin{minipage}[t]{1\linewidth}
			\centering
			\includegraphics[width=6.6in]{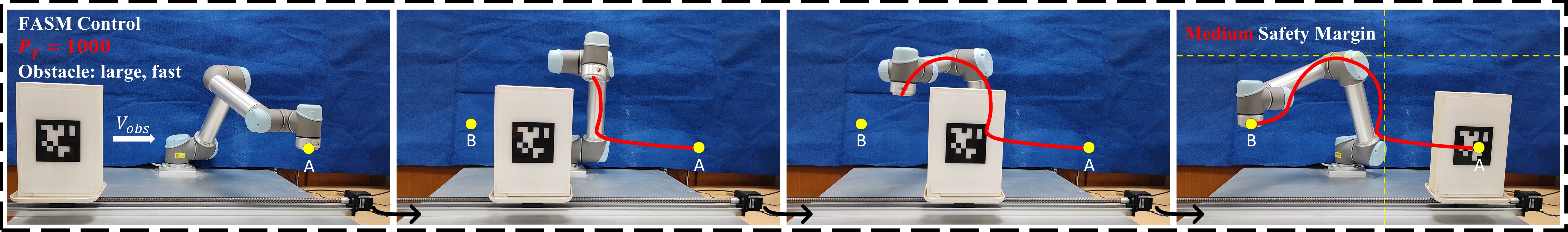}
			%\caption{fig1}
		\end{minipage}%
	}%	\setcounter{subfigure}{0}
	\vspace{-2.1mm}
	\setcounter{subfigure}{0}	
	\subfigure{
		\begin{minipage}[t]{1\linewidth}
			\centering
			\includegraphics[width=6.6in]{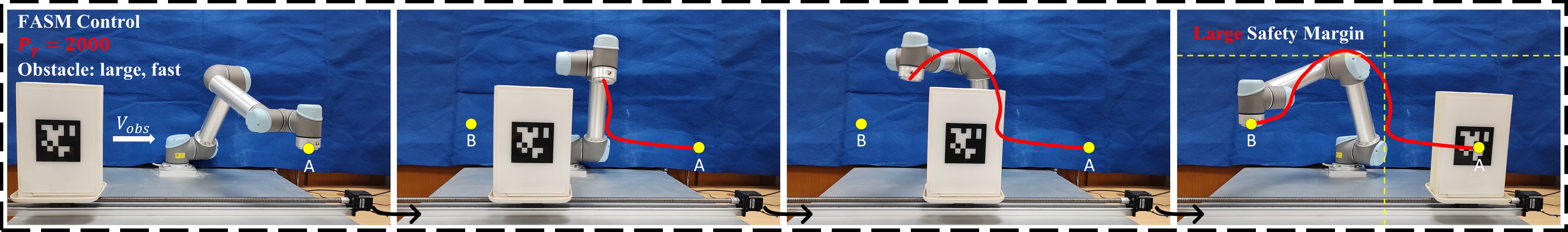}
			%\caption{fig1}
		\end{minipage}%
	}%	\setcounter{subfigure}{0}
	% Ìí¼ÓÌâ×¢£¬¼´¶ÔÕâ¸öÍ¼Æ¬µÄËµÃ÷
	\caption{Experiment 3: Frame-by-frame plots in the scenario of the fast-moving large obstacle under different parameters. (Only the trajectory from point A to point B is shown.)}
	\label{fig:result_include1}
\end{figure*}
\begin{figure*}[t!]
	\centering
	%µÚÒ»ÐÐÍ¼Æ¬Õ¹Ê¾
	\subfigure{
		\begin{minipage}[t]{0.23\linewidth}
			\centering
			\includegraphics[width=1.8in]{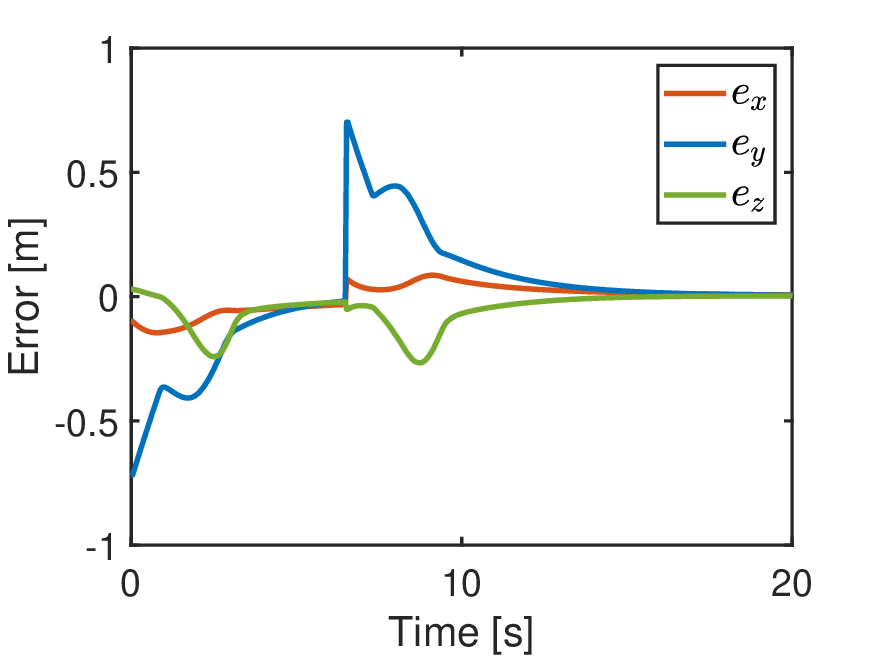}
			%\caption{fig1}
		\end{minipage}%
	}%	\setcounter{subfigure}{0}
	\subfigure{
		\begin{minipage}[t]{0.23\linewidth}
			\centering
			\includegraphics[width=1.8in]{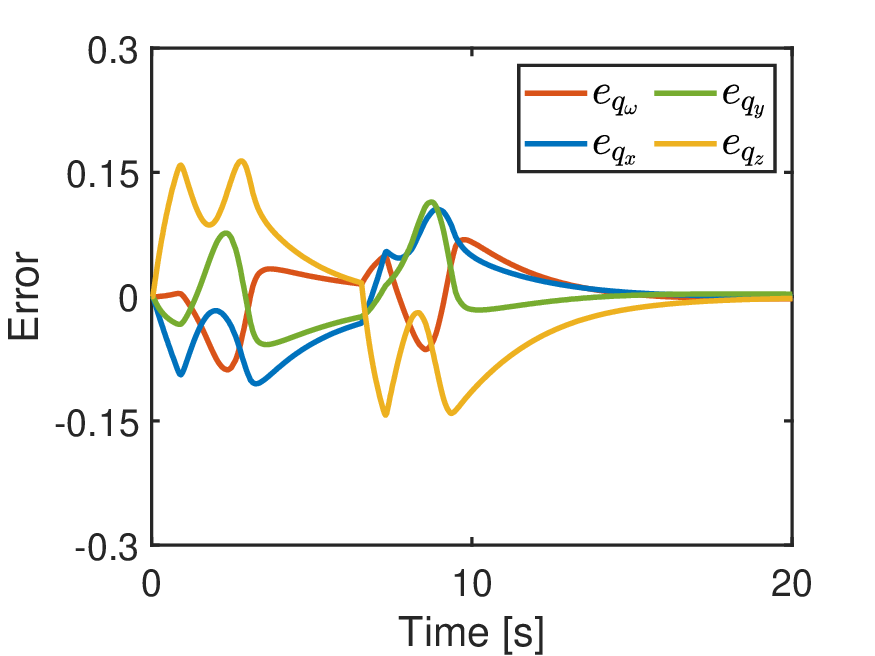}
			%\caption{fig1}
		\end{minipage}%
	}%
	\subfigure{
		\begin{minipage}[t]{0.23\linewidth}
			\centering
			\includegraphics[width=1.8in]{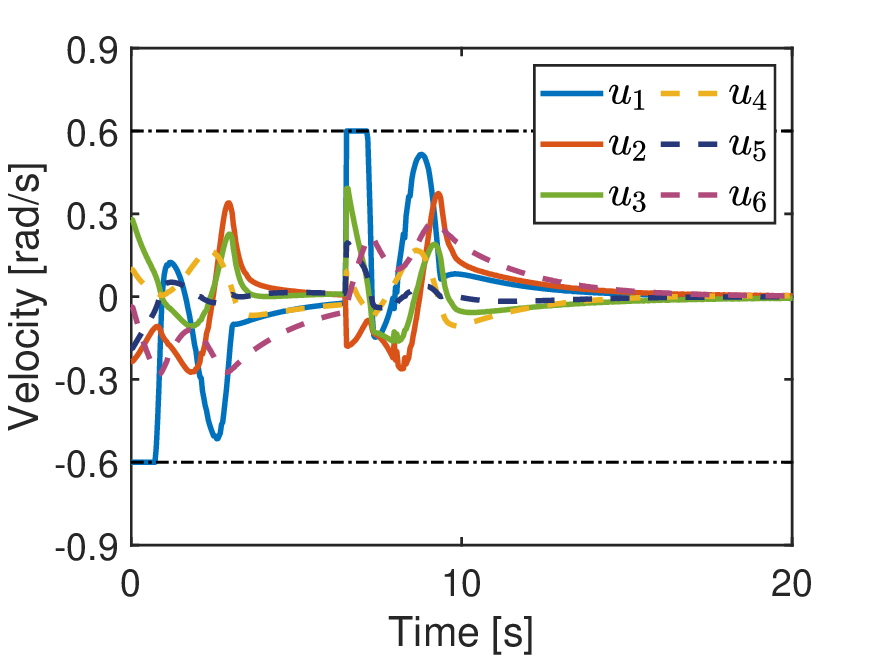}
			%\caption{fig1}
		\end{minipage}%
	}%
	\subfigure{
		\begin{minipage}[t]{0.25\linewidth}
			\centering
			\includegraphics[width=1.85in]{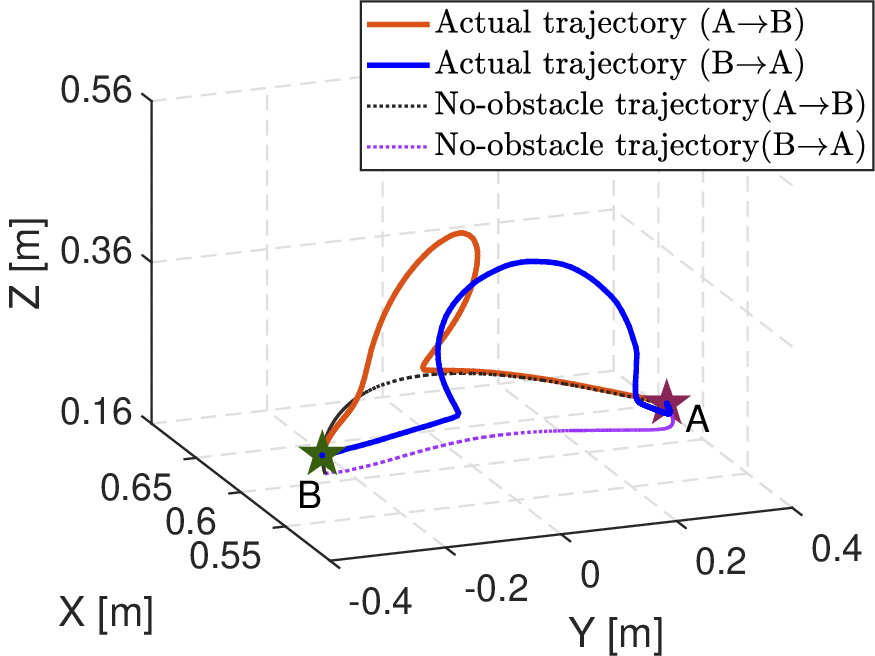}
			%\caption{fig1}
		\end{minipage}%
	}%
	\vspace{-3.5mm}
	\setcounter{subfigure}{0}	
	%	% µÚ¶þÐÐÍ¼Æ¬Õ¹Ê¾
	\subfigure{
		\begin{minipage}[t]{0.23\linewidth}
			\centering
			\includegraphics[width=1.8in]{big_error_xyz_p1000.eps}
		\end{minipage}%
	}%
	\subfigure{
		\begin{minipage}[t]{0.23\linewidth}
			\centering
			\includegraphics[width=1.8in]{big_error_quan_p1000.eps}
		\end{minipage}%
	}%
	\subfigure{
		\begin{minipage}[t]{0.23\linewidth}
			\centering
			\includegraphics[width=1.8in]{big_u_p1000.eps}
		\end{minipage}%
	}%
	\subfigure{
		\begin{minipage}[t]{0.25\linewidth}
			\centering
			\includegraphics[width=1.85in]{big_trajectory_p1000.eps}
			%\caption{fig1}
		\end{minipage}%
	}%
	\vspace{-3.5mm}
	\setcounter{subfigure}{0}	
	% µÚÈýÐÐÍ¼Æ¬Õ¹Ê¾
	\subfigure[Position errors]{
		\begin{minipage}[t]{0.23\linewidth}
			\centering
			\includegraphics[width=1.8in]{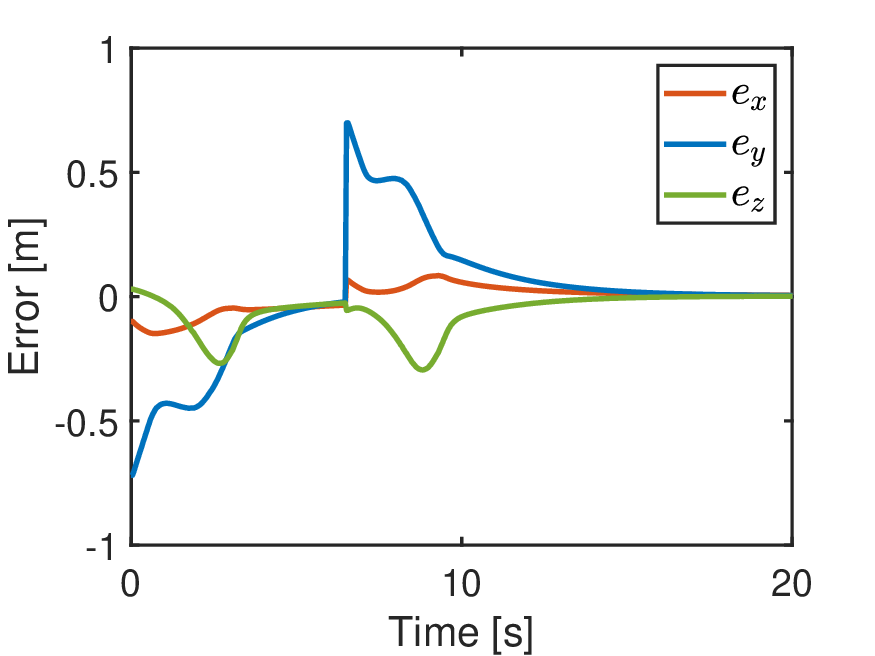}
		\end{minipage}%
	}%
	\subfigure[Quaternion errors]{
		\begin{minipage}[t]{0.23\linewidth}
			\centering
			\includegraphics[width=1.8in]{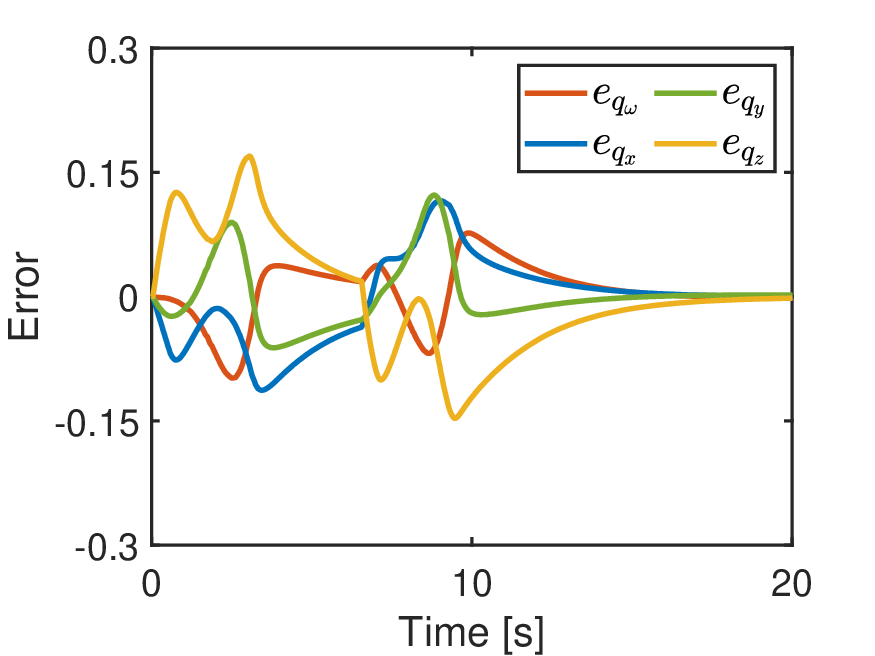}
		\end{minipage}%
	}%
	\subfigure[Joint velocities]{
		\begin{minipage}[t]{0.23\linewidth}
			\centering
			\includegraphics[width=1.8in]{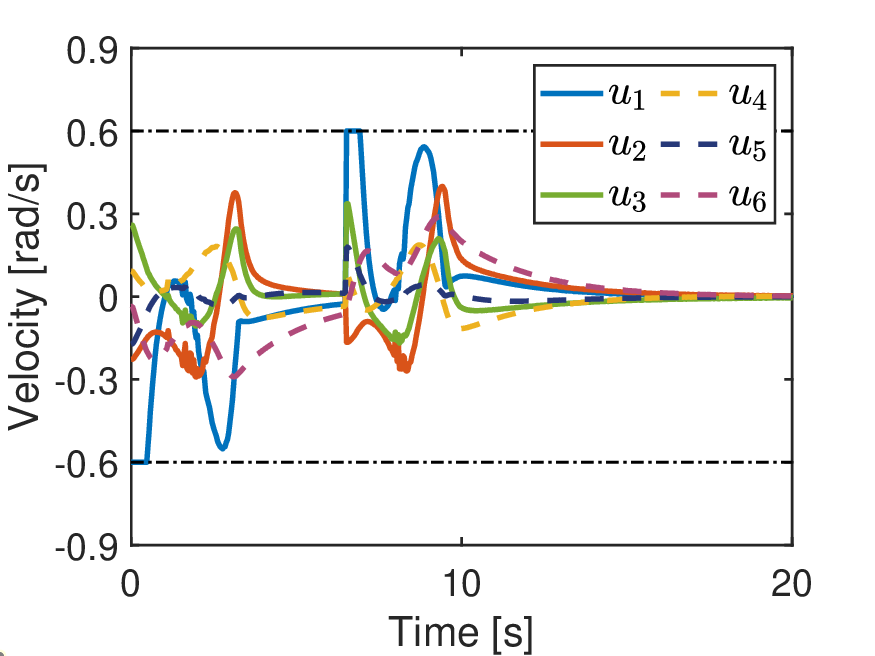}
		\end{minipage}%
	}%
	\subfigure[Trajectories of the end-effector]{
		\begin{minipage}[t]{0.25\linewidth}
			\centering
			\includegraphics[width=1.85in]{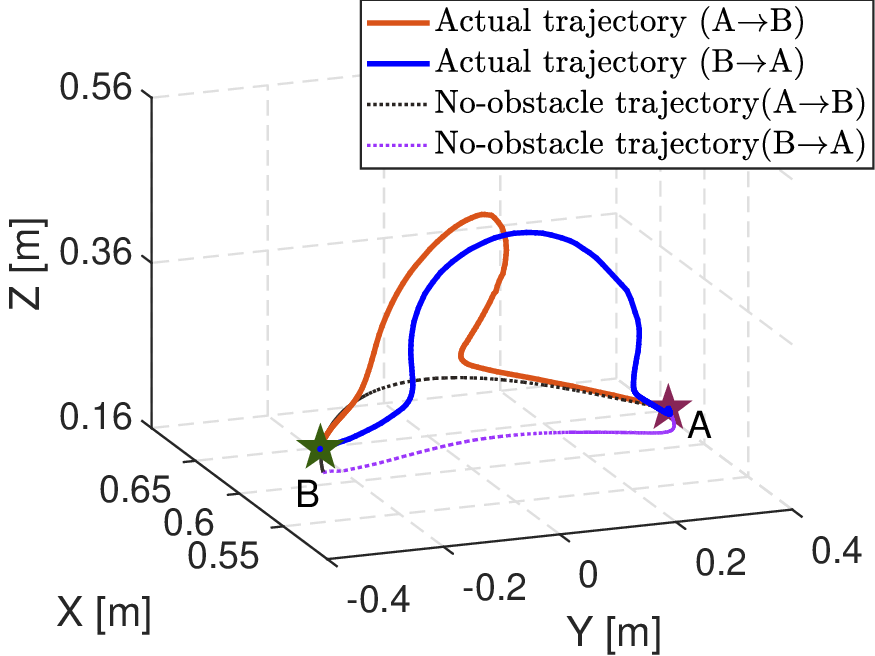}
			%\caption{fig1}
		\end{minipage}%
	}%
	% Ìí¼ÓÌâ×¢£¬¼´¶ÔÕâ¸öÍ¼Æ¬µÄËµÃ÷
	\caption{Experiment 3: Position errors, quaternion errors, joint velocities and end-effector trajectories of the proposed FASM control in the scenario of the fast-moving large obstacle under different values of $P_\gamma$ (top: $P_\gamma=200$; middle: $P_\gamma=1000$; and bottom: $P_\gamma=2000$).}
	\label{fig:result_include1}
\end{figure*}
\vspace{-1mm}
\section{Conclusion}
To summarize, the FASM control approach has been proposed in this paper to offer flexibility and active safety for robot manipulators in dynamic environments with moving obstacles. A GPIO was first introduced to estimate the dynamic information of the obstacles, enabling short-term prediction of obstacle motion. Subsequently, the estimation errors were quantified to refine the design of safety distances. Next, the discrete-time CBFs and estimation information were employed to formulate the flexible CBFSC. Following that, the MPC philosophy was applied, integrating the flexible CBFSC as safety constraints into the optimization problem. Notably, the decay rates of the flexible CBFSC were dynamically optimized to enhance flexibility throughout the dynamic obstacle avoidance process. Finally, the experimental results in various scenarios have demonstrated that the proposed FASM control effectively addresses the challenges of dynamic obstacle avoidance, reference tracking and constraints satisfaction for robot manipulators in dynamic environments. %It significantly enhanced performance in terms of flexibility and active safety throughout the dynamic obstacle avoidance process.

\vfill


\begin{thebibliography}{1}
\bibliographystyle{IEEEtran}
\bibitem{application1} M. Safeea, P. Neto, and R. Bearee, ``On-Line Collision Avoidance for Collaborative Robot Manipulators by Adjusting Off-Line Generated Paths: An Industrial Use Case," \textit{Robotics and Autonomous Systems}, vol. 119, pp. 278-288, Jul. 2019.	
\bibitem{application2} J. Burgner-Kahrs, D. C. Rucker, and H. Choset, ``Continuum Robots for Medical Applications: A Survey," \textit{IEEE Transactions on Robotics}, vol. 31, no. 6, pp. 1261-1280, Dec. 2015.	
\bibitem{application3} C. Weng, Q. Yuan, F. Suarez-Ruiz, and I. Chen, ``A Telemanipulation-Based Human-Robot Collaboration Method to Teach Aerospace Masking Skills," \textit{IEEE Transactions on Industrial Informatics}, vol. 16, no. 5, pp. 3076-3084, May 2020.
\bibitem{robot_uncer} B. Xiao, and S. Yin, ``Exponential Tracking Control of Robotic Manipulators With Uncertain Dynamics and Kinematics," \textit{IEEE Transactions on Industrial Informatics}, vol. 15, no. 2, pp. 689-698, Feb. 2019.

\bibitem{intro1} F. Ferraguti et al.,  ``Safety and Efficiency in Robotics: The Control Barrier Functions Approach," \textit{IEEE Robotics and Automation Magazine}, vol. 29, no. 3, pp. 139-151, Sept. 2022.

%\bibitem{plan1} Y. Dai, S. Yu, Y. Yan, and X. Yu, ``An EKF-Based Fast Tube MPC Scheme for Moving Target Tracking of a Redundant Underwater Vehicle-Manipulator System," \textit{IEEE/ASME Transactions on Mechatronics}, vol. 24, no. 6, pp. 2803-2814, Dec. 2019.
%Ma H, Li C, Liu J, et al, ``Enhance connectivity of promising regions for sampling-based path planning," \textit{IEEE Transactions on Automation Science and Engineering}, vol. 20, no. 3, pp. 1997-2010, Jul. 2023.
%\bibitem{plan2} T. Gold, A. Volz, and K. Graichen, ``Model Predictive Interaction Control for Robotic Manipulation Tasks," \textit{IEEE Transactions on Robotics}, vol. 39, no. 1, pp. 76-89, Feb. 2023.
\bibitem{plan1} W. Zhang, H. Cheng , L. Hao, et al, ``An Obstacle Avoidance Algorithm for Robot Manipulators Based on Decision-Making Force," \textit{Robotics and Computer-Integrated Manufacturing}, vol. 71, Oct. 2021.

\bibitem{RRT1} S. M. LaValle, ``Rapidly-Exploring Random Trees: A New Tool for Path Planning," \textit{Research Report}, 1998.
\bibitem{RRT2} L. Jiang, S. Liu, Y. Cui, and H. Jiang, ``Path Planning for Robotic Manipulator in Complex Multi-Obstacle Environment Based on Improved$\_$RRT," \textit{IEEE/ASME Transactions on Mechatronics}, vol. 27, no. 6, pp. 4774-4785, Dec. 2022.

\bibitem{search_based1} P. E. Hart, N. J. Nilsson, and B. Raphael, ``A Formal Basis for the Heuristic Determination of Minimum Cost Paths," \textit{IEEE Transactions on Systems Science and Cybernetics}, vol. 4, no. 2, pp. 100-107, Jul. 1968.
\bibitem{search_based2} S. Liu, K. Mohta, N. Atanasov, and V. Kumar, ``Search-Based Motion Planning for Aggressive Flight in SE(3)," \textit{IEEE Robotics and Automation Letters}, vol. 3, no. 3, pp. 2439-2446, Jul. 2018.

%\bibitem{H1} S. X., Yang, and  M., Meng, "An Efficient Neural Network Approach to Dynamic Robot Motion Planning," \textit{Neural networks}, vol. 13, no. 2, pp. 143-148, Mar. 2000. 
\bibitem{H2} Z. Zhang, L. Zheng, J. Yu, Y. Li, and Z. Yu, ``Three Recurrent Neural Networks and Three Numerical Methods for Solving a Repetitive Motion Planning Scheme of Redundant Robot Manipulators," \textit{IEEE/ASME Transactions on Mechatronics}, vol. 22, no. 3, pp. 1423-1434, Jun. 2017.
\bibitem{H3} Y. Hu, and S. X. Yang, ``A Knowledge Based Genetic Algorithm for Path Planning of A Mobile Robot," in \textit{IEEE International Conference on Robotics and Automation}, ICRA, New Orleans, LA, USA, 2004, pp. 4350-4355, vol. 5.
%\bibitem{H4} A. Bakdi, A. Hentout,  H. Boutami, A. Maoudj, O. Hachour, and B. Bouzouia, ``Optimal path planning and execution for mobile robots using genetic algorithm and adaptive fuzzy-logic control," \textit{Robotics and Autonomous Systems}, vol. 89, pp. 95-109, Mar. 2017.
%\bibitem{H5} T. T. Mac, C. Copot, D. T. Tran, and R.D. Keyser, ``Heuristic approaches in robot path planning: A survey," \textit{Robotics and Autonomous Systems}, vol. 86, pp. 13-28, Dec. 2016.

%\bibitem{PF} C. Thorpe, and L. Matthies, ``Path Relaxation: Path Planning for a Mobile Robot," in \textit{OCEANS}, Washington, DC, USA, 1984, pp. 576-581.

\bibitem{Op1} X. Zhang, A. Liniger, and F. Borrelli, ``Optimization-Based Collision Avoidance," \textit{IEEE Transactions on Control Systems Technology}, vol. 29, no. 3, pp. 972-983, May 2021.
%\bibitem{Op2} J. Schulman, Y. Duan, J. Ho et al., ``Motion planning with sequential convex optimization and convex collision checking," \textit{The International Journal of Robotics Research}, vol. 33, no. 9, pp. 1251-1270, Aug. 2014.

%Tanzmeister, G., Wollherr, D., and Buss, M, `` Grid-based multi-road-course estimation using motion planning," \textit{IEEE Transactions on Vehicular Technology}, vol. 65, no. 4, pp. 1924-1935, Apr. 2016.
%\bibitem{plan3} Huang, Y., Ding, H., Zhang, Y., et al, ``A motion planning and tracking framework for autonomous vehicles based on artificial potential field elaborated resistance network approach," \textit{IEEE Transactions on Industrial Electronics}, pp. 1376-1386, Feb. 2020.
%\bibitem{plan4} Ma, J., Li, X., Liang, W., and Tan, K. K, ``Parameter space optimization towards constrained controller design with application to tray indexing," \textit{IEEE Transactions on Industrial Electronics}, vol. 67, no. 7, pp. 5575-5585, Jul. 2020.
%\bibitem{sampling} J. Vannoy, and J. Xiao, ``Real-Time Adaptive Motion Planning (RAMP) of Mobile Manipulators in Dynamic Environments With Unforeseen Changes," \textit{IEEE Transactions on Robotics}, vol. 24, no. 5, pp. 1199-1212, Oct. 2008.

%\bibitem{u} A. Singletary, P. Nilsson, T. Gurriet, and A. D. Ames, ``Online Active Safety for Robotic Manipulators," in \textit{2019 IEEE/RSJ International Conference on Intelligent Robots and Systems (IROS)}, Macau, China, pp. 173-178, 2019.	
\bibitem{APF1} O. Khatib, ``Real-Time Obstacle Avoidance for Manipulators and Mobile Robots," \textit{The International Journal of Robotics Research}, vol. 5, no. 1, pp. 90-98, Mar. 1986.

\bibitem{APF2} Y. Tian, X. Zhu, D. Meng, X. Wang, and B. Liang, ``An Overall Configuration Planning Method of Continuum Hyper-Redundant Manipulators Based on Improved Artificial Potential Field Method," \textit{IEEE Robotics and Automation Letters}, vol. 6, no. 3, pp. 4867-4874, Jul. 2021.

\bibitem{mpc_safe3} T. Zhu, J. Mao, L. Han, C. Zhang, and J. Yang, ``Real-Time Dynamic Obstacle Avoidance for Robot Manipulators Based on Cascaded Nonlinear MPC With Artificial Potential Field," \textit{IEEE Transactions on Industrial Electronics}, doi: 10.1109/TIE.2023.3306405. 

\bibitem{fuzzy1} A. Hentout, A. Maoudj, and M. Aouache, ``A Review of the Literature on Fuzzy-Logic Approaches for Collision-Free Path Planning of Manipulator Robots," \textit{Artificial Intelligence Review}, vol. 56, no. 4, pp. 3369-3444, Sept. 2023.

\bibitem{fuzzy2} E. A. Merchan-Cruz, and A. S. Morris, ``Fuzzy-GA-Based Trajectory Planner for Robot Manipulators Sharing A Common Workspace," \textit{IEEE Transactions on Robotics}, vol. 22, no. 4, pp. 613-624, Aug. 2006.

%\bibitem{nn1} R. Glasius, A. Komoda, and S. C. Gielen, ``Neural Network Dynamics for Path Planning and Obstacle Avoidance," \textit{Neural Networks}, vol. 8, no. 1, pp. 125-133, Feb. 1995)

\bibitem{learn1} B. Sangiovanni, A. Rendiniello, G. P. Incremona, A. Ferrara, and M. Piastra, ``Deep Reinforcement Learning for Collision Avoidance of Robotic Manipulators," in \textit{2018 European Control Conference (ECC)}, Limassol, Cyprus, 2018, pp. 2063-2068.

\bibitem{learn2} P. Chen, J. Pei, W. Lu, and M. Li, ``A Deep Reinforcement Learning Based Method for Real-Time Path Planning and Dynamic Obstacle Avoidance," \textit{Neurocomputing}, vol. 497, pp. 64-75, May 2022.

\bibitem{QP1} D. Guo, and Y. Zhang, ``A New Inequality-Based Obstacle-Avoidance MVN Scheme and Its Application to Redundant Robot Manipulators," \textit{IEEE Transactions on Systems, Man, and Cybernetics, Part C (Applications and Reviews)}, vol. 42, no. 6, pp. 1326-1340, Nov. 2012.

\bibitem{qpnn} Y. Zhang, and J. Wang, ``Obstacle Avoidance for Kinematically Redundant Manipulators Using A Dual Neural Network," \textit{IEEE Transactions on Systems, Man, and Cybernetics, Part B (Cybernetics)}, vol. 34, no. 1, pp. 752-759, Feb. 2004.

\bibitem{QP2}Z. Xu, X. Zhou, H. Wu, X. Li, and S. Li, ``Motion Planning of Manipulators for Simultaneous Obstacle Avoidance and Target Tracking: An RNN Approach With Guaranteed Performance," \textit{IEEE Transactions on Industrial Electronics}, vol. 69, no. 4, pp. 3887-3897, Apr. 2022.

%\bibitem{mpc1} D. Q. Mayne, ``Model Predictive Control: Recent Developments and Future Promise," \textit{Automatica}, vol. 50, no. 12, pp. 2967-2986, Nov. 2014.	
%\bibitem{mpc2} Z. Jin, D. Qin, A. Liu, W. Zhang, and L. Yu, ``Model Predictive Variable Impedance Control of Manipulators for Adaptive Precision-Compliance Trade off," \textit{IEEE/ASME Transactions on Mechatronics}, vol. 28, no. 2, pp. 1174-1186, Apr. 2023.
%\bibitem{mpcbook} G. P. Incremona, A. Ferrara, and L. Magni, ``MPC for Robot Manipulators With Integral Sliding Modes Generation," \textit{IEEE/ASME Transactions on Mechatronics}, vol. 22, no. 3, pp. 1299-1307, Jun. 2017.
\bibitem{mpc3} J. Liu, J. Yang, S. Li, and X. Wang, ``Single-Loop Robust Model Predictive Speed Regulation of PMSM Based on Exogenous Signal Preview," \textit{IEEE Transactions on Industrial Electronics}, vol. 70, no. 12, pp. 12719-12729, Dec. 2023.	

%\bibitem{ddc1} A. H. Khan, S. Li and X. Luo, ``Obstacle Avoidance and Tracking Control of Redundant Robotic Manipulator: An RNN-Based Metaheuristic Approach," \textit{IEEE Transactions on Industrial Informatics}, vol. 16, no. 7, pp. 4670-4680, Jul. 2020.
%\bibitem{ddc2} Y. Wen, and P. Pagilla, ``Path-Constrained and Collision-Free Optimal Trajectory Planning for Robot Manipulators," \textit{IEEE Transactions on Automation Science and Engineering}, vol. 20, no. 2, pp. 763-774, Apr. 2023.


%\bibitem{mpc_safe1} A. Oleinikov, S. Kusdavletov, A. Shintemirov, and M. Rubagotti, ``Safety-Aware Nonlinear Model Predictive Control for Physical Human-Robot Interaction," \textit{IEEE Robotics and Automation Letters}, vol. 6, no. 3, pp. 5665-5672, Jul. 2021.
%\bibitem{TandC} Z. Li, J. Deng, R. Lu, Y. Xu, J. Bai, and C. Su, ``Trajectory-Tracking Control of Mobile Robot Systems Incorporating Neural-Dynamic Optimized Model Predictive Approach," \textit{IEEE Transactions on Systems, Man, and Cybernetics: Systems}, vol. 46, no. 6, pp. 740-749, Jun. 2016.
\bibitem{mpc_safe1} J. Nubert, J. Kohler, V. Berenz, F. Allgower, and S. Trimpe, ``Safe and Fast Tracking on A Robot Manipulator: Robust MPC and Neural Network Control," \textit{IEEE Robotics and Automation Letters}, vol. 5, no. 2, pp. 3050-3057, Apr. 2020.  
\bibitem{mpc_safe2} A. Zube, ``Cartesian Nonlinear Model Predictive Control of Redundant Manipulators Considering Obstacles," in \textit{2015 IEEE International Conference on Industrial Technology (ICIT)}, Seville, Spain, 2015, pp. 137-142.
\bibitem{mpc_safe4} A. S. Sathya, J. Gillis, G. Pipeleers, and J. Swevers, ``Real-time Robot Arm Motion Planning and Control With Nonlinear Model Predictive Control Using Augmented Lagrangian on A First-Order Solver," in \textit{2020 European Control Conference (ECC)}, St. Petersburg, Russia, 2020, pp. 507-512.
\bibitem{mpc_safe5} M. Rubagotti, B. Sangiovanni, A. Nurbayeva, G. P. Incremona, A. Ferrara, and A. Shintemirov, ``Shared Control of Robot Manipulators With Obstacle Avoidance: A Deep Reinforcement Learning Approach," \textit{IEEE Control Systems Magazine}, vol. 43, no. 1, pp. 44-63, Feb. 2023.

%\bibitem{mpc_safe4} Bing Z, Mavrichev A, Shen S, et al, ``Safety Guaranteed Manipulation Based on Reinforcement Learning Planner and Model Predictive Control Actor," \textit{arXiv preprint arXiv}:2304.09119, 2023.	

\bibitem{mpccbf1} J. Zeng, B. Zhang, and K. Sreenath, ``Safety-Critical Model Predictive Control With Discrete-Time Control Barrier Function," in \textit{2021 American Control Conference (ACC)}, New Orleans, LA, USA, 2021, pp. 3882-3889.

%\bibitem{mpccbf2} Vahidi-Moghaddam A, Chen K, Zhang K, et al, ``A Unified Framework for Online Data-Driven Predictive Control with Robust Safety Guarantees," \textit{arXiv preprint arXiv}:2306.17270, 2023.


\bibitem{CBF1} 	A. D. Ames, X. Xu, J. W. Grizzle, and P. Tabuada, ``Control Barrier Function Based Quadratic Programs for Safety Critical Systems," \textit{IEEE Transactions on Automatic Control}, vol. 62, no. 8, pp. 3861-3876, Aug. 2017.	
%\bibitem{CBF2} X. Tan, W. S. Cortez, and D. V. Dimarogonas, ``High-Order Barrier Functions: Robustness, Safety, and Performance-Critical Control," \textit{IEEE Transactions on Automatic Control}, vol. 67, no. 6, pp. 3021-3028, Jun. 2022.	
\bibitem{Dis_CBF} A. Agrawal, and K. Sreenath, ``Discrete Control Barrier Functions for Safety-Critical Control of Discrete Systems With Application to Bipedal Robot Navigation," \textit{Robotics: Science and Systems}, vol. 13, 2017.
\bibitem{Dis_CBF2} Y. Xiong, D. Zhai, M. Tavakoli, and Y. Xia, ``Discrete-Time Control Barrier Function: High-Order Case and Adaptive Case," \textit{IEEE Transactions on Cybernetics}, vol. 53, no. 5, pp. 3231-3239, May 2023.

\bibitem{CBF_rob1} A. Singletary, S. Kolathaya, and A. D. Ames, ``Safety-Critical Kinematic Control of Robotic Systems," \textit{IEEE Control Systems Letters}, vol. 6, pp. 139-144, Jan. 2021.
\bibitem{CBF_rob2} M. A. Murtaza, S. Aguilera, M. Waqas, and S. Hutchinson, ``Safety Compliant Control for Robotic Manipulator With Task and Input Constraints," \textit{IEEE Robotics and Automation Letters}, vol. 7, no. 4, pp. 10659-10664, Oct. 2022.
%\bibitem{CBF_rob3} J. Lin, D. Zhai, Y. Xiong, and Y. Xia, ``Safety Control for UR-Type Robotic Manipulators via High-Order Control Barrier Functions and Analytical Inverse Kinematics," \textit{IEEE Transactions on Industrial Electronics}, vol. 71, no. 6, pp. 6150-6160, Jun. 2024.
\bibitem{CBF_rob5} Y. Tang, X. Chu, J. Huang, and K. W. Samuel Au, ``Learning-Based MPC With Safety Filter for Constrained Deformable Linear Object Manipulation," \textit{IEEE Robotics and Automation Letters}, vol. 9, no. 3, pp. 2877-2884, Mar. 2024.
%\bibitem{CBF_rob4} D. Zhang, M. Van, S. Mcllvanna, Y. Sun, and S. McLoone, ``Adaptive Safety-Critical Control With Uncertainty Estimation for Human-Robot Collaboration," \textit{IEEE Transactions on Automation Science and Engineering}, doi: 10.1109/TASE.2023.3320873.

\bibitem{CBF_fea1} J. Zeng, Z. Li, and K. Sreenath, ``Enhancing Feasibility and Safety of Nonlinear Model Predictive Control With Discrete-Time Control Barrier Functions," in \textit{2021 60th IEEE Conference on Decision and Control (CDC)}, Austin, TX, USA, 2021, pp. 6137-6144.
\bibitem{CBF_fea2} J. Zeng, B. Zhang, Z. Li, and K. Sreenath, ``Safety-Critical Control Using Optimal-Decay Control Barrier Function With Guaranteed Point-Wise Feasibility," in \textit{2021 American Control Conference (ACC)}, New Orleans, LA, USA, 2021, pp. 3856-3863.
\bibitem{CBF_fea3} Z. Lu, K. Feng, J. Xu, H. Chen, and Y. Lou,  ``Robot Safe Planning in Dynamic Environments Based on Model Predictive Control Using Control Barrier Function." \textit{arXiv preprint arXiv:2404.05952}, 2024.
\bibitem{CBF_fea4} R. Periotto, M. Ferizbegovic, F. S. Barbosa, and R. C. Sundin, ``MPC-CBF With Adaptive Safety Margins for Safety-Critical Teleoperation over Imperfect Network Connections," \textit{arXiv preprint arXiv:2403.18650}, 2024.
\end{thebibliography}
\end{document}